\documentclass{article}
\usepackage{iclr2027_conference,times}

\usepackage{amsmath,amsfonts,bm}

\def\eqref#1{equation~\ref{#1}}

\def\1{\bm{1}}

\DeclareMathAlphabet{\mathsfit}{\encodingdefault}{\sfdefault}{m}{sl}
\SetMathAlphabet{\mathsfit}{bold}{\encodingdefault}{\sfdefault}{bx}{n}

\def\eqref#1{(\ref{#1})}

\usepackage{url}
\usepackage{booktabs}
\usepackage{graphicx}
\usepackage{float}
\usepackage{capt-of}
\usepackage{amsmath,amssymb}
\usepackage[colorlinks=true,linkcolor=blue,citecolor=blue,urlcolor=blue]{hyperref}
\usepackage{siunitx}
\graphicspath{{figs/}}

\renewcommand{\arraystretch}{0.95}
\usepackage{microtype}

\newcommand{\Dold}{\mathcal{D}_{\mathrm{old}}}
\newcommand{\Dnew}{\mathcal{D}_{\mathrm{new}}}
\newcommand{\thbase}{\theta_{\mathrm{base}}}

\title{A Deafening Silence: Catastrophic Forgetting\\
Lives in the Output Embeddings of Tokens\\
the Data Never Speaks}
\author{Jonghyun Han$^{1}$, Younghoon Song$^{2}$, Jongyoul Park$^{1,*}$ \\
$^{1}$Seoul National University of Science and Technology \\
$^{2}$Korea Institute of Land \& Infrastructure Safety Technology \\
$^{*}$Corresponding author: \texttt{jongyoul@seoultech.ac.kr}}

\iclrfinalcopy

\begin{document}
\maketitle
\lhead{Preprint}

\begin{abstract}
Continual pre-training and fine-tuning in Large Language Models (LLMs) inevitably induce catastrophic forgetting, typically mitigated by replay using often-inaccessible original data. In this data-free regime, we analyze where forgetting occurs and why. Systematic parameter freezing across five settings up to 1.4B reveals that forgetting concentrates selectively in the output embeddings of tokens rarely seen in the new corpus, whereas the same $\sqrt{\hat{v}}$ band of the body is inert and new learning resides elsewhere. This localization is governed by the vocabulary deficiency of the corpus rather than the training mode, allowing pre-retraining risk ranking from token counts alone within a fixed base model. Mechanistically, absent tokens receive persistent one-sided softmax gradients that Adam's second-moment ($\sqrt{\hat{v}}$) normalization amplifies into full-sized updates. We therefore propose an intervention: raising Adam's $\epsilon$ exclusively for the output projection during training. Across eight settings spanning 160M to 12B parameters and four model families, this removes 39.4\% to 67.9\% of forgetting across all seven stable configurations without degrading target learning or requiring per-model tuning. The defense combines additively or better with replay (79.8\% on Qwen/Korean) and rescues released-head LoRA from a 23-fold forgetting surge. Because post-hoc editing of the drifted rows recovers under 5\% of forgetting, the intervention must operate during training. Our findings indicate that a single-line optimizer adjustment may serve as the primary defense against catastrophic forgetting where the corpus starves the vocabulary.
\end{abstract}

\section{Introduction}
\label{sec:intro}

\begin{figure}[t]
\centering
\includegraphics[width=\textwidth]{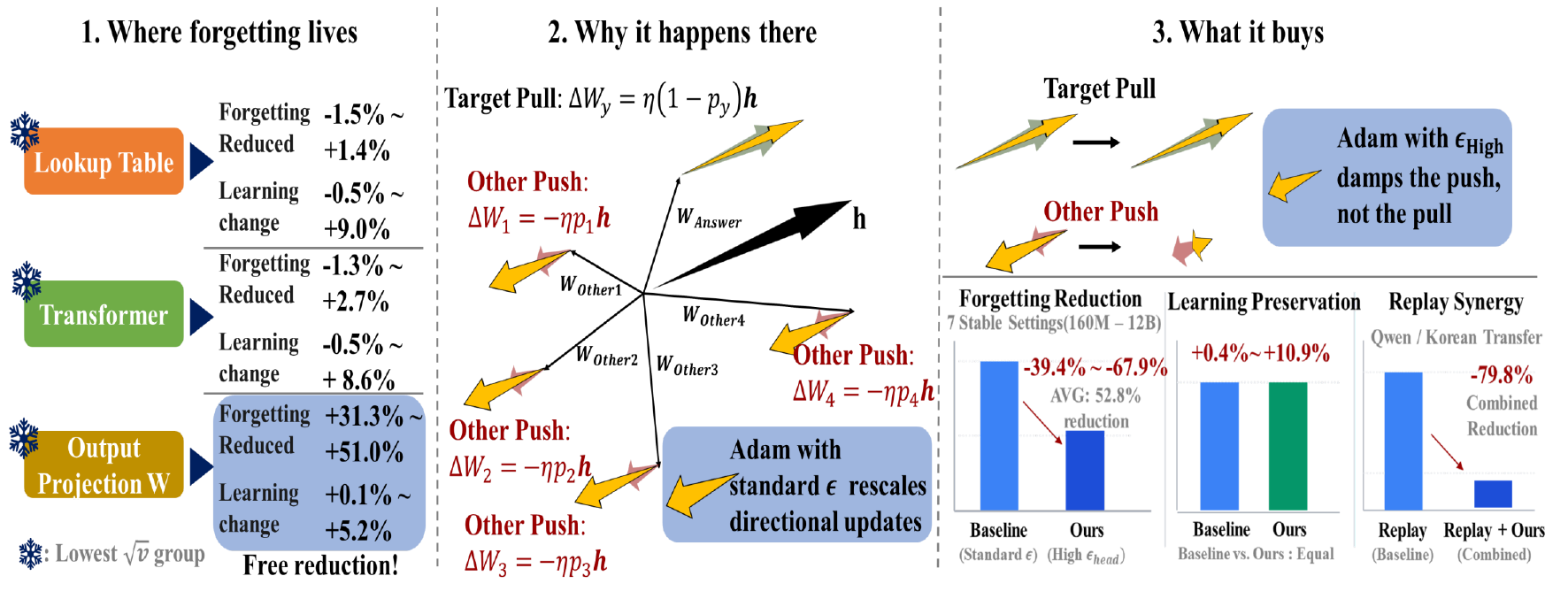}
\caption{Overview: where the forgetting lives, why it lives there, and what acting on it buys.}
\label{fig:teaser}
\end{figure}

Adapting pre-trained foundation models via continual pre-training or fine-tuning inevitably induces catastrophic forgetting~\citep{kirkpatrick2017ewc}. While replay mitigates this by mixing original pre-training data~\citep{ibrahim2024simple}, such data is rarely accessible. In this data-free regime, we causally localize where forgetting occurs and target that exact site to prevent it (Figure~\ref{fig:teaser}).

Prior research on forgetting divides into mitigation and localization. Beyond replay~\citep{ibrahim2024simple,hickok2025scalable}, data-free defenses (including parameter selection~\citep{chen2025mofo}, low-rank adapters~\citep{hu2022lora,biderman2024lora}, and orthogonal subspace projections~\citep{saha2021gpm,wang2023orthogonal}) indiscriminately constrain aggregate parameter drift, sacrificing plasticity for retention. Conversely, localization studies report conflicting sites across early attention, lower layers, and FFNs~\citep{gu2020investigating,zheng2025spurious,mechanistic2026forgetting}, primarily due to correlational metrics like weight displacement. Although theory suggests forgetting concentrates in low-rank output modes~\citep{hidekel2026lowrank}, the specific tensors holding these modes remain unidentified.

Instead of unguided regularization, we intervene directly: freezing parameter groups to their pre-trained values isolates each group's causal share of forgetting. Varying sites, second-moment ($\sqrt{\hat{v}}$) bands, and corpora pinpoints forgetting to under-exposed output embeddings. Because these embeddings drift when Adam assigns full step sizes to minute softmax gradients, raising $\epsilon$ exclusively on this tensor suppresses drift without impeding learning.

Our contributions are as follows:
\begin{itemize}
\item \textbf{Causal localization}: By training-time freezing, we show that forgetting concentrates in output embeddings of rare tokens (31--51\% at five stable settings; the body in the same band remains $<10\%$). This site is governed by corpus vocabulary deficiency, enabling pre-retraining risk ranking simply by counting tokens.
\item \textbf{Underlying mechanism}: Absent token embeddings receive persistent one-sided softmax gradients, which Adam's $\sqrt{\hat{v}}$ normalization amplifies into full-sized updates that default $\epsilon$ fails to curb.
\item \textbf{Output-scoped $\epsilon$}: Elevating $\epsilon$ exclusively on the output projection eliminates 39.4--67.9\% of forgetting across seven stable settings of an eight-setting grid, 160M to 12B parameters and four model families without degrading target learning or requiring original data. It composes additively with replay and prevents catastrophic forgetting in released-head LoRA.
\item \textbf{Boundary conditions}: Post-hoc editing of the drifted rows recovers $<5\%$ of forgetting, so the intervention must operate during training; when forgetting shifts to the body under in-distribution tuning, damping it costs learning at one of the two cells we measure, because body coordinates conflate drift with new knowledge (Section~\ref{sec:bodyeps}).
\end{itemize}

\section{Related Work}
\label{sec:related}

\textbf{Replay and defenses without original data.}
Continual pre-training defenses typically rely on replay and learning rate schedules~\citep{ibrahim2024simple,hickok2025scalable}. Without original data, existing methods restrict parameter drift via weight regularization, parameter selection~\citep{chen2025mofo}, or low-rank adapters~\citep{biderman2024lora}, yet these blindly constrain aggregate movement without separating forgetting from learning. Indeed, releasing the output head alone under fixed-rank LoRA multiplies forgetting 23-fold on a tied model and 16-fold on an untied one (Section~\ref{sec:baselines}, Appendix~\ref{app:lora}). We focus on data-free settings and view replay as complementary.

\textbf{Localizing forgetting.}
Reported sites of forgetting conflict across early attention, lower layers, and FFNs~\citep{gu2020investigating,zheng2025spurious,mechanistic2026forgetting}, largely due to correlational metrics like weight displacement. While output freezing mitigates degradation~\citep{gu2020investigating}, the specific rows responsible and their causal mechanisms remained unexplained. Unlike class-incremental learning where output bias stems from task imbalance~\citep{wu2019large,masana2023class}, continual pre-training lacks explicit label skew. Using training-time freezing, we show that corpus vocabulary composition causally dictates where forgetting resides.

\textbf{The geometry of rare tokens and remedies.}
Persistent negative softmax gradients push rare token embeddings into a narrow cone~\citep{gao2019representation,yu2022rare}. While low-rank dynamics are well-documented~\citep{aghajanyan2021intrinsic,hidekel2026lowrank}, existing subspace methods~\citep{saha2021gpm,wang2023orthogonal} require past data, and ASR token freezing~\citep{kwok2024continual} relies on binary heuristics. We show this drift directly carries forgetting, identify Adam as the amplifier and $\epsilon$ as the brake, and measure rank directly from drift. Unlike post-hoc editing~\citep{ilharco2023editing,wortsman2022robust} or optimizer tuning for model scaling~\citep{choi2019empirical,wortsman2023small,yang2023tensor,everett2024scaling}, ours operates during training.

\section{Preliminaries}

\subsection{Problem Setting}
\label{sec:problem}

\textbf{Setting and constraint.}
We adapt a pre-trained foundation model $\thbase$ (trained on an inaccessible corpus $\Dold$) to a new corpus $\Dnew$ via continual pre-training or fine-tuning, aiming to acquire new domain knowledge without catastrophic forgetting. We assume zero access to $\Dold$, requiring strictly data-free adaptation, although combinations with replay are evaluated in Section~\ref{sec:main}.

\textbf{Notation.}
Consider a transformer with vocabulary size $V$ and hidden dimension $d$, trained with AdamW. Given context representation $h$, the unembedding matrix $W \in \mathbb{R}^{V \times d}$ produces logits $z_t = h^{\top} w_t$, where $w_t \in \mathbb{R}^{d}$ is row $t$ of $W$, the output embedding of token $t$. Let $c(t)$ denote the target occurrence count of token $t$ in $\Dnew$; tokens with $c(t) < K$ ($K = 1{,}000$) are designated as under-exposed. \emph{Vocabulary deficiency} is the occurrence-weighted fraction of old-domain evaluation positions whose target token is under-exposed in $\Dnew$, counted over the four-domain pool of Section~\ref{sec:cells} minus the injected domain, the wider $\mathcal{E}$ of Table~\ref{tab:domain}. Computed simply by tokenizing both corpora with the base tokenizer without retraining, it separates tasks into domain injection (high deficiency) and in-distribution tuning (low deficiency).

\subsection{Background}
\label{sec:math}

\textbf{Softmax gradients and Adam's $\bm{\epsilon}$.}
Given context $h$, output probabilities are $p = \mathrm{softmax}(Wh)$ with target loss $L = -\log p_y$. The gradient for output embedding $w_t$ is
\begin{equation}
\frac{\partial L}{\partial w_t} = \frac{\partial L}{\partial z_t} \frac{\partial z_t}{\partial w_t} = \bigl(p_t - \mathbf{1}[t = y]\bigr)\, h.
\label{eq:grad}
\end{equation}
Target embeddings are pulled toward $h$, whereas unselected tokens are pushed along $-p_t h$, persistently depressing their logits. In contrast, input embeddings update only upon occurrence.

AdamW pairs Adam's adaptive step with a decoupled decay~\citep{kingma2015adam,loshchilov2019adamw}:
\begin{equation}
\Delta\theta = -\,\eta\left(\frac{\hat{m}}{\sqrt{\hat{v}} + \epsilon} + \lambda\theta\right).
\label{eq:adam}
\end{equation}
For sign-consistent gradients, $\hat{m} \approx g$ and $\sqrt{\hat{v}} \approx |g|$, pushing the effective step size toward $\eta$ regardless of gradient scale $|g|$. Oscillating gradients cancel in $\hat{m}$, so update magnitude reflects directional consistency rather than gradient norm. The denominator $\epsilon$ acts as a floor: when $\sqrt{\hat{v}} \lesssim \epsilon$, it scales updates by $\sqrt{\hat{v}}/(\sqrt{\hat{v}} + \epsilon)$, selectively braking parameters with small $\sqrt{\hat{v}}$ (default: $\epsilon = \num{1e-8}$).

\subsection{Experimental Setup}
\label{sec:cells}

\textbf{The grid of settings.}
Each setting pairs a foundation model with a new domain (Table~\ref{tab:cells}). The Pythia suite (160M--12B) enables controlled scaling comparisons with an openly verifiable $\Dold$~\citep{biderman2023pythia}. Qwen2.5-0.5B and TinyLlama-1.1B provide tied and untied embeddings, respectively, while OLMo-2-1B represents a modern post-2024 architecture. Target domains span Korean, code, math, and instruction tuning (Alpaca and KoAlpaca), with Korean and Alpaca at opposite extremes of vocabulary deficiency.

\begin{table}[b]
\caption{The grid of settings. Korean is the primary injection domain across all base models; other domains vary vocabulary deficiency at fixed models (Section~\ref{sec:spectrum}). All settings run replay-free for 40M tokens across three seeds (12.6M for instruction tuning; Table~\ref{tab:module}) unless noted.}
\label{tab:cells}
\centering
\scriptsize
\begin{tabular}{lll}
\toprule
Base model (architecture) & Primary domain & Additional domains \\
\midrule
Pythia 160M--12B (GPT-NeoX)    & Korean & math, code, Alpaca, KoAlpaca (410M and 1.4B only) \\
Qwen2.5-0.5B (tied)            & Korean & math, code, Alpaca, KoAlpaca \\
TinyLlama-1.1B (Llama, untied) & Korean & -- \\
OLMo-2-1B (olmo2, untied)      & Korean & -- \\
\bottomrule
\end{tabular}
\end{table}

\textbf{Training procedure.}
All models are trained with AdamW. Continual pre-training uses a 40M token budget (Table~\ref{tab:cells}), instruction tuning uses 12.6M, and budget checks extend to 160M and 400M tokens; all comparisons are strictly budget-matched. Interventions halt (freeze) or dampen (raise $\epsilon$) updates to specific parameter groups throughout training. We separately sweep learning rates on a log grid ($10^{-5}$ to $10^{-3}$ for pre-training, $10^{-6}$ to $3 \times 10^{-4}$ for instruction tuning), selecting the peak on target learning and breaking ties within seed variance toward lower rates (Table~\ref{tab:lrcurve}, Table~\ref{tab:locstats}).

\textbf{Evaluation corpora.}
Validation losses are evaluated in nats (cross-entropy per token) on fixed held-out corpora. The old-domain set $\mathcal{E}$ excludes the adaptation target: English and code for Korean/math injections; English and math for code; English, code, and math for instruction tuning. Reductions are strictly compared within one $\mathcal{E}$; the domain grid reads each row on the pool minus that row's injection, so its corpora are wider and differ by row (Table~\ref{tab:domain}). These corpora sample $\Dold$ directly for Pythia and OLMo, and act as proxies for closed base models.

\textbf{Metrics.}
Target learning is evaluated on $\Dnew$ validation loss, while forgetting averages loss increases across all domains in $\mathcal{E}$:
{\small
\begin{align}
\mathrm{learn} &= L_{\mathrm{base}}(\Dnew) - L_{\mathrm{after}}(\Dnew), &
\mathrm{forget} &= \frac{1}{|\mathcal{E}|} \sum_{d \in \mathcal{E}} \bigl[L_{\mathrm{after}}(d) - L_{\mathrm{base}}(d)\bigr], \label{eq:learn} \\
\mathrm{reduction} &= 1 - \mathrm{forget}_{\mathrm{int}} / \mathrm{forget}_{\mathrm{ref}}, &
\mathrm{change} &= (\mathrm{learn}_{\mathrm{int}} - \mathrm{learn}_{\mathrm{ref}}) / \mathrm{learn}_{\mathrm{ref}}, \label{eq:reduction}
\end{align}}
where $L_{\mathrm{base}}$ and $L_{\mathrm{after}}$ denote pre- and post-retraining losses, and subscripts $\mathrm{ref}$ and $\mathrm{int}$ indicate reference and intervention runs at tuned learning rates. Positive $\mathrm{learn}$ and $\mathrm{forget}$ denote capability gains and incurred losses, while $\mathrm{reduction}$ and $\mathrm{change}$ are relative to the baseline.

\textbf{Seeds and verdicts.}
We report three-seed averages unless noted; arms with seed disagreement were expanded to six or more. An intervention is judged to preserve learning if $\mathrm{change}$ remains within baseline seed variation: $(\max - \min) / \mathrm{mean}$ of reference $\mathrm{learn}$ across seeds. This is 2--4\% at most settings but far wider where the intervention arm is bimodal, reaching 18.9\% at Pythia-410M on Korean.

\section{Where Forgetting Lives, and Why There}
\label{sec:diagnosis}

\begin{figure}[t]
\centering
\includegraphics[width=\textwidth]{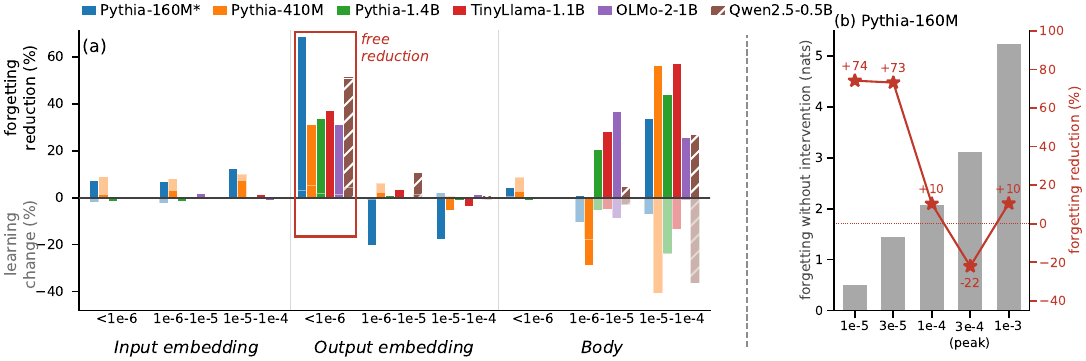}
\caption{(a) Causal share by (site, $\sqrt{\hat{v}}$ band): bars denote forgetting reduction (up) and learning change (down) from freezing one group (Korean, replay-free). Hatched: Qwen (tied) at \num{1e-4}. Pythia-160M ($*$) is at \num{1e-5} due to peak-rate collapse. (b) Pythia-160M across rates, measured with output-scoped $\epsilon$ rather than freezing (Table~\ref{tab:rate160m}). Full grid: Table~\ref{tab:locfull}.}
\label{fig:loc}
\end{figure}

\subsection{Freezing One Group at a Time Locates the Forgetting}
\label{sec:loc}

\textbf{Measurement protocol: causal share.}
We define a parameter group's causal share as the forgetting reduction obtained by freezing only that group to its pre-trained values during training. Unlike correlational drift metrics, this directly tests necessity: a parameter shift holds causal share only if preventing it reduces forgetting.

We partition three components (input embeddings, output embeddings, body) across three $\sqrt{\hat{v}}$ bands ($0$--\num{1e-6}, \num{1e-6}--\num{1e-5}, \num{1e-5}--\num{1e-4}), freezing one pair at a time using fixed baseline masks. To separate location from parameter count, we include mass-matched controls freezing equivalent body parameters. This grid costs eleven arms per setting, so it runs at the six settings up to 1.4B rather than the eight of Section~\ref{sec:main}, on replay-free Korean adaptation at baseline peak learning rates, except Pythia-160M which collapses at its peak and is evaluated at $1 \times 10^{-5}$ (detailed in (iv)). Tied Qwen treats embeddings as one component.

\textbf{Result: forgetting and learning live in different places.}
Causal shares reveal four primary findings:

\textbf{(i) Forgetting localizes to the bottom output band.} Freezing output embeddings with $\sqrt{\hat{v}} < \num{1e-6}$ eliminates 31.3--51.0\% of forgetting across five settings, with target learning shifting by $+0.1\%$ to $+5.2\%$ (no degradation). Freezing input embeddings in the same band yields negligible reduction ($-1.5\%$ to $+1.4\%$; four untied settings), confirming that output embeddings drive forgetting.

\textbf{(ii) Location governs forgetting, not low $\sqrt{\hat{v}}$ alone.} The low-$\sqrt{\hat{v}}$ body band holds negligible share ($-1.3\%$ to $+2.7\%$). Although this band covers 74--86\% of output parameters versus only 0.3--1.4\% in the body, mass-matched body freezing yields only 0.4--12.3\% reduction across five stable settings (Table~\ref{tab:massmatch}; Pythia-160M is the exception at 28.5\%, detailed in (iv)). Selection matters within the table too: at Qwen, random output rows remove 12.1\% at a $4.5\%$ learning cost against 42.7\% for free from the lowest-$\sqrt{\hat{v}}$ coordinates (Appendix~\ref{app:equiv}). Disabling weight decay leaves forgetting and reduction essentially unchanged in Qwen and shifts Pythia-410M only within its seed spread (Appendix~\ref{app:tables}, Table~\ref{tab:wd}). Thus, forgetting is causally governed by row location, not parameter count or regularization artifacts.

\textbf{(iii) Learning localizes to upper body bands.} Freezing upper body bands eliminates 8.8--40.7\% of target learning (the costlier upper band per setting), while freezing output embeddings leaves learning intact. Because forgetting and learning occupy disjoint parameter regions, halting output drift imposes no plasticity penalty.

\textbf{(iv) Total collapse overrides localized mitigation.} At its peak rate ($3 \times 10^{-4}$), Pythia-160M suffers total collapse (3.1 nats baseline forgetting), where freezing fails. Lowering the learning rate restores mitigation. Measured with output-scoped $\epsilon$ rather than with freezing, reductions reach 74.2\% ($1 \times 10^{-5}$) and 73.2\% ($3 \times 10^{-5}$) where baseline forgetting stays below 1.5 nats, against 10.3\% at 2.1 nats and $-22.1\%$ at 3.1 nats (Figure~\ref{fig:loc}b, Table~\ref{tab:rate160m}). Rerunning the freezing grid at $1 \times 10^{-5}$ recovers the standard pattern (Table~\ref{tab:locfull}, column $\dagger$): the bottom output band achieves a 68.8\% share ($+3.1\%$ learning), input bands remain below 13\%, the bottom body band holds 4.6\%, and the cost of freezing reappears in the upper body bands. Breakdown occurs only under extreme baseline forgetting, not due to model scale. Summary ranges in (i)--(iii) exclude this outlier.

\begin{table}[t]
\begin{minipage}[t]{0.40\textwidth}
\setlength{\tabcolsep}{2pt}
\centering
\scriptsize
\captionof{table}{Count-matched site contrast: parameters frozen (M) and the forgetting reduction (\%) from freezing an equal mass at each site's low end (matched to the lowest output band). Each row divides by one reference trained at that setting's peak rate. n/a: tied embeddings.}
\label{tab:massmatch}
\resizebox{\linewidth}{!}{%
\begin{tabular}{lrrrr}
\toprule
Setting & \shortstack[c]{Params\\frozen (M)} & \shortstack[c]{Output\\emb.} & \shortstack[c]{Input\\emb.} & Body \\
\midrule
Pythia-160M     & 30.7  & $-6.2$ & 9.8    & 28.5 \\
Pythia-410M     & 38.7  & 31.4   & 2.6    & 11.2 \\
Pythia-1.4B     & 80.6  & 33.7   & $-0.5$ & 0.4 \\
TinyLlama-1.1B  & 48.2  & 37.0   & 0.4    & 2.9 \\
OLMo-2-1B       & 176.6 & 31.3   & $-1.0$ & 4.4 \\
Qwen2.5-0.5B    & 110.0 & 51.0   & n/a    & 12.3 \\
\bottomrule
\end{tabular}}

\end{minipage}\hfill
\begin{minipage}[t]{0.575\textwidth}
\centering
\scriptsize
\captionof{table}{Reference forgetting strictly follows vocabulary deficiency, while reduction does not (output $\epsilon = \num{1e-4}$; references three seeds, interventions two, 410M/ko six per arm). Each cell is at its own peak rate, so rows are not rate-matched (Appendix~\ref{app:tables}). Learning changes, where negative, stay within each reference's seed spread (18.9\% for 410M/ko). Here $\mathcal{E}$ is the four-domain pool minus each row's injected domain, wider than the $\mathcal{E}$ of Table~\ref{tab:main} and not identical across rows, so cells are comparable with neither.}
\label{tab:domain}
\resizebox{\linewidth}{!}{%
\begin{tabular}{llrrrr}
\toprule
Base model & Injected & \shortstack[c]{Vocab.\\deficiency (\%)} & \shortstack[c]{Ref.\ forget\\(nats)} & \shortstack[c]{Reduction\\(\%)} & \shortstack[c]{Learning\\change (\%)} \\
\midrule
Pythia-410M  & math & 33.9 & 0.0431 & 36.4 & $-3.7$ \\
             & code & 55.9 & 0.0803 & 55.7 & $+0.4$ \\
             & ko   & 63.2 & 0.6565 & 31.0 & $-5.4$ \\
\midrule
Pythia-1.4B  & math & 33.9 & 0.0560 & 24.6 & $-0.3$ \\
             & code & 55.9 & 0.0832 & 53.1 & $+4.5$ \\
             & ko   & 63.2 & 0.5121 & 40.5 & $+2.2$ \\
\midrule
Qwen2.5-0.5B & math & 50.6 & 0.0258 & 67.8 & $+10.0$ \\
             & code & 52.5 & 0.1806 & 72.7 & $+18.4$ \\
             & ko   & 61.3 & 0.6255 & 62.8 & $+5.7$ \\
\bottomrule
\end{tabular}}

\end{minipage}
\end{table}

\subsection{The Data, Not the Training Mode, Decides That Location}
\label{sec:spectrum}

\textbf{Vocabulary deficiency sets how much is forgotten.}
Holding the model, the budget and the four-domain pool fixed while varying the target domain across math, code, and Korean (Table~\ref{tab:domain}; each cell keeps its own peak rate and drops its own injection from $\mathcal{E}$, so rows match on neither), baseline forgetting strictly follows vocabulary deficiency in all three models: 0.043, 0.080, 0.657 nats (410M); 0.056, 0.083, 0.512 nats (1.4B); and 0.026, 0.181, 0.626 nats (Qwen). Because vocabulary deficiency is computed simply by tokenizing both corpora beforehand, it correctly ranks injections within each model, although it does not predict cross-model absolute magnitudes (Spearman rank correlation 0.60; Appendix~\ref{app:preflight}). Relative mitigation is decoupled from deficiency, with code yielding the largest reduction. At 410M on Korean, six seeds yield a wide reduction spread (4.6\% to 46.6\%, mean 31.0\%) coupled with learning change: ordered by reduction, the same seeds order almost exactly by learning change ($-32.9\%$ to $+16.8\%$), so low-mitigation runs reflect a failure to take hold, not a retention-plasticity trade-off.

\textbf{Separating training mode from data.}
Comparing Alpaca and KoAlpaca under instruction fine-tuning varies the data distribution at a fixed task format (Table~\ref{tab:module}), with the caveat that only Pythia-410M is rate-matched across the pair. In-distribution Alpaca concentrates causal share in MLPs, with output embeddings in single digits for untied models (11.6\% in tied Qwen). KoAlpaca introduces novel vocabulary, shifting share to under-exposed output embedding rows. While freezing whole output modules in untied models yields negligible reduction ($-3.7\%$, $-1.6\%$), output-specific $\epsilon$ on low-$\sqrt{\hat{v}}$ rows eliminates 29.8--43.3\% of forgetting without sacrificing target learning. The apparent 46.9\% module freeze share in Qwen is an artifact of parameter tying. Furthermore, freezing MLPs on KoAlpaca reduces forgetting by 8.2--47.8\% but sacrifices 16.3--26.3\% of target learning, confirming that body freezing severely impairs acquisition.

\textbf{What the attention column does not settle.}
Freezing attention on KoAlpaca removes 42.7\% of forgetting at 410M (against 29.8\% for output $\epsilon$) and 30.2\% at 1.4B (against 33.1\%), at negligible learning cost. Because causal shares do not partition across unnormalized modules, an unmatched module freeze that holds a large share of the weights cannot be directly contrasted with a one-group optimizer change. Our claim is specific: the share recoverable by a row-level rule shifts to output embeddings when new vocabulary is introduced. Whether attention holds an independent share remains open (Appendix~\ref{app:module}).

\begin{table}[t]
\caption{Instruction fine-tuning (replay-free, 12.6M tokens): forgetting reduction (learning change). Freezes hold modules at base; $\epsilon$ slows low-$\sqrt{\hat{v}}$ output rows ($\dagger$ freezes Qwen's tied table). Each cell is at its own peak rate; only 410M is rate-matched across the Alpaca/KoAlpaca pair (Appendix~\ref{app:tables}).}
\label{tab:module}
\centering
\scriptsize
\renewcommand{\arraystretch}{1.45}%
\begin{tabular}{lcccccc}
\toprule
 & \multicolumn{2}{c}{Pythia-410M} & \multicolumn{2}{c}{Pythia-1.4B} & \multicolumn{2}{c}{Qwen2.5-0.5B} \\
\cmidrule(lr){2-3}\cmidrule(lr){4-5}\cmidrule(lr){6-7}
 & Alpaca & KoAlpaca & Alpaca & KoAlpaca & Alpaca & KoAlpaca \\
\midrule
Attention freeze          & 16.0 ($-1.0$) & 42.7 ($-0.7$) & 7.0 ($-0.5$) & 30.2 ($+0.4$) & 17.2 ($-1.5$) & 15.5 ($+0.1$) \\
MLP freeze                & 27.9 ($-5.8$) & 8.2 ($-23.1$) & 39.7 ($-3.7$) & 47.8 ($-16.3$) & 44.8 ($-15.4$) & 26.1 ($-26.3$) \\
Output-embedding freeze   & 2.7 ($-1.2$) & $-3.7$ ($+3.0$) & 0.3 ($+0.1$) & $-1.6$ ($+0.5$) & 11.6 ($-3.1$)$^{\dagger}$ & 46.9 ($-3.9$)$^{\dagger}$ \\
Input-embedding freeze    & 1.6 ($-0.0$) & 6.6 ($+1.1$) & 0.8 ($+0.2$) & $-1.4$ ($+0.5$) & n/a & n/a \\
Output-scoped $\epsilon$  & 14.6 ($-0.1$) & 29.8 ($+6.2$) & 5.6 ($-0.3$) & 33.1 ($+0.9$) & 6.0 ($-1.0$) & 43.3 ($-1.4$) \\
\bottomrule
\end{tabular}
\end{table}

\subsection{Why Those Embeddings Move: Softmax and Adam}
\label{sec:mech}

\textbf{The softmax selects embeddings, Adam enlarges their step.}
Output embeddings absent as targets in $\Dnew$ receive only the one-sided gradient $p_t h$ at each token position, which drives updates along $-p_t h$. Because context $h$ is shared without opposing pulls, these gradients are small, sign-consistent, and directionally aligned. By Eq.~\eqref{eq:adam}, Adam maximally amplifies these updates: $\hat{m}$ and $\sqrt{\hat{v}}$ scale down together, driving effective step sizes toward the full learning rate. In contrast, input embeddings update only on explicit occurrences. In five untied settings, 34.8--51.2\% of input parameters retain $\sqrt{\hat{v}} = 0$ post-training (identically 36.59\% across the three Pythia models sharing a tokenizer; Table~\ref{tab:v_distribution}), whereas no output coordinate remains zero across all six settings (Table~\ref{tab:vsite}). Drift thus concentrates in output rows accumulating persistent negative gradients everywhere.

\textbf{Empirical evidence: travel distance across $\bm{\epsilon}$.}
Travel distances over 120 steps confirm this mechanism. Under default $\epsilon = 10^{-8}$, parameter displacement stays within one order of magnitude across bands despite disparate gradient scales. Directional consistency (net displacement over total path length) peaks in the bottom band at 0.67--0.90, indicating unidirectional drift. Raising $\epsilon$ to $10^{-4}$ introduces sharp selectivity, damping displacement by 48.0--189.2$\times$ in the lowest band versus only 1.4--1.8$\times$ in the highest (Table~\ref{tab:travel}, which bands the whole model and so cannot attribute damping to a site). Default $\epsilon$ remains inactive because only 3.5--15.4\% of parameters in Pythia (5.0\% in Qwen) fall below $10^{-8}$ (Table~\ref{tab:v_distribution}). Because median output coordinates sit two to three orders of magnitude below median body coordinates (Appendix~\ref{app:vsite}), elevating $\epsilon$ selectively brakes output drift. Softmax selects which embeddings drift, Adam dictates how far, and default $\epsilon$ fails to brake.

\section{The Intervention and Its Boundaries}
\label{sec:method}

\subsection{The Intervention and Its Equivalents}
\label{sec:intervention}

\textbf{The intervention.}
We set $\epsilon = \num{1e-4}$ exclusively for the output projection in optimizer parameter groups, introducing zero compute or memory overhead without requiring explicit target token lists. Under Eq.~\eqref{eq:adam}, $\epsilon$ scales each step by $\sqrt{\hat{v}} / (\sqrt{\hat{v}} + \epsilon)$: at $\epsilon = \num{1e-4}$ a coordinate at the output median (\num{1.1e-8} to \num{7.8e-8}) keeps under a thousandth of its step, while one at the body median (\num{5.5e-6} to \num{4.2e-5}) keeps a twentieth to a third. The gap between the sites (Appendix~\ref{app:vsite}), not a hand-placed threshold, supplies the selectivity; sensitivity is in Table~\ref{tab:epsval}.

\textbf{Equivalents.}
The key mechanism is arresting under-exposed rows rather than $\epsilon$ itself. Two other instruments achieve equivalent results: freezing rows with $c(t) < 1{,}000$ at base values, and per-step gating updating rows only when tokens appear in the batch. Although operating differently, all three agree closely in forgetting reduction across five settings (Appendix~\ref{app:equiv}, Figure~\ref{fig:equiv} and Table~\ref{tab:equiv}). We adopt $\epsilon$ because it requires no explicit token tracking.

\textbf{Controls.}
Scaling down the embedding learning rate traces the same frontier, and at a matched 96.6\% of the output embedding (under 1\% replay) random and count-based masks retain similarly but differ sharply in learning cost: mass buys retention, the criterion decides its price (Appendix~\ref{app:equiv}).

\textbf{Choosing the scope.}
The intervention site is predictable prior to retraining. If baseline forgetting is non-positive, no intervention is needed. Otherwise, tokenize the candidate corpus with the base tokenizer to compute the share of old-domain positions with target counts $c(t) < K = 1{,}000$. A high fraction indicates output-dominated forgetting, warranting output-specific $\epsilon$. A low fraction indicates body-dominated forgetting, where freezing harms target learning (Section~\ref{sec:spectrum}, Table~\ref{tab:module}), making replay or learning rate scaling preferable. This heuristic strictly matched empirical forgetting ranks at the three models of Table~\ref{tab:domain} and at four earlier (checkpoint, corpus) comparisons; quantitative analyses and failed extensions appear in Appendix~\ref{app:preflight}.

\subsection{Main Result: Free Reduction at Eight Settings}
\label{sec:main}

\textbf{Generality across architectures and scales.}
We evaluate output-specific $\epsilon$ across eight settings under Korean domain injection (Table~\ref{tab:main}). Target learning is preserved at three seeds ($+0.4\%$ to $+10.9\%$; the bimodal 410M arm reads $-7.0\%$ at six, inside its 18.9\% spread), with forgetting reduction spanning 39.4\% to 67.9\% across seven standard models. Pythia-160M collapses at its peak rate (Section~\ref{sec:loc}), so its row reflects $3 \times 10^{-5}$ and is excluded from summary ranges. Baseline forgetting in the Pythia suite decreases monotonically with tokens-per-parameter exposure, falling from 3.1 nats at 160M's peak rate to 0.10 nats at 12B (Tables~\ref{tab:rate160m} and~\ref{tab:main}). Under a fixed budget, exposure and model size co-vary, so this column alone cannot separate them.

\textbf{The largest models are quiet because they are underfed, not because they are large.}
Reference forgetting at 6.9B and 12B is approximately 0.1 nats, which might appear negligible. However, extending the token budget demonstrates that forgetting at scale is deferred rather than absent. At each run's own fixed rate to 160M tokens ($3 \times 10^{-5}$ at 6.9B, $10^{-5}$ at 12B), baseline forgetting grows from 0.238 to 0.305 nats in 6.9B and from 0.044 to 0.076 nats in 12B between the 40M and 160M checkpoints. Rather than decaying, the reduction rises from 53.9\% to 62.3\% at 6.9B and from 76.2\% to 84.1\% at 12B, while learning changes stay within $+2.2\%$ and $+0.6\%$ across all checkpoints (Table~\ref{tab:budget}). These are checkpoints inside 160M-token runs whose schedule has not yet decayed, so the ladder compares one run against itself. Likewise, Pythia-410M trained on code across a tenfold budget (400M tokens) incurs 1.27 nats of baseline forgetting, where the intervention eliminates 56.2\% while improving target learning by 10.1\% (Appendix~\ref{app:code}). Budget scale is therefore not a flattering confound; it is the axis along which the defense strengthens.

\begin{table}[t]
\caption{Output-only $\epsilon$ (Korean, replay-free, 40M tokens, peak lr; 3 seeds, 2 for 12B/OLMo baselines). $^*$160M is at \num{3e-5} due to peak collapse (Figure~\ref{fig:loc}b, Table~\ref{tab:rate160m}). The 410M arm is bimodal across seeds: at six seeds of the same configuration it reads 32.7\% and $-7.0\%$ (Table~\ref{tab:wd}).}
\label{tab:main}
\centering
\scriptsize
\resizebox{0.85\linewidth}{!}{%
\begin{tabular}{lrrrrr}
\toprule
Setting & Tokens/param & Ref.\ forget (nats) & Interv.\ forget & Reduction (\%) & Learning change (\%) \\
\midrule
Pythia-160M$^{*}$ & 0.2469 & 1.448 & 0.388 & 73.2 & $+9.9$ \\
Pythia-410M     & 0.0988 & 0.664 & 0.373 & 43.8    & $+10.9$ \\
Qwen2.5-0.5B    & 0.0810 & 0.629 & 0.202 & 67.9    & $+5.4$ \\
TinyLlama-1.1B  & 0.0364 & 0.361 & 0.187 & 48.1    & $+0.4$ \\
Pythia-1.4B     & 0.0284 & 0.473 & 0.261 & 44.8    & $+2.2$ \\
OLMo-2-1B       & 0.0269 & 1.137 & 0.689 & 39.4    & $+1.8$ \\
Pythia-6.9B     & 0.0058 & 0.109 & 0.044 & 60.2    & $+0.8$ \\
Pythia-12B      & 0.0034 & 0.101 & 0.035 & 65.3    & $+1.1$ \\
\bottomrule
\end{tabular}}
\end{table}

\textbf{Operating regime and robustness.}
Mitigation operates within a distinct learning-rate regime: forgetting reduction holds steady across the low-rate plateau, then collapses at excessive rates where target learning fails alongside it, so $\epsilon$ cannot replace rate tuning, and at a 40M budget that plateau is narrowest at 6.9B and 12B. Because the optimal rate is invariant to the intervention, a single standard sweep suffices; nor does the method require per-model $\epsilon$ tuning. Appendix~\ref{app:lrcurve} reads the rate grid (Table~\ref{tab:lrcurve}) and the $\epsilon$ sweep (Table~\ref{tab:epsval}). These benefits survive capability benchmarks and tenfold budget expansion (Appendix~\ref{app:checks}); math and code injections follow in Table~\ref{tab:domain}.

\textbf{Interaction with experience replay.}
When replay is viable, both defenses combine effectively. On Qwen/Korean, the combined reduction (79.8\%) matches the independence prediction (78.0\%) within 2 percentage points, with $\epsilon$'s individual share remaining invariant to replay (69.5\% vs.\ 72.1\%, Table~\ref{tab:replay}). In OLMo, the combination becomes super-additive (74.4\% vs.\ 64.8\% predicted). Across both settings, the workflow is identical: activate the zero-cost optimizer modification first, then deploy replay to curb residual forgetting at token expense.

\begin{table}[t]
\begin{minipage}[t]{0.36\textwidth}
\centering
\scriptsize
\captionof{table}{Replay $\times$ output $\epsilon$ (nats; reduction \%; 3 seeds). Separate campaigns, Qwen at lr \num{3e-5} and 40M, OLMo at lr \num{1e-4} and 60M. Learning stays in spread.}
\label{tab:replay}
\resizebox{\linewidth}{!}{%
\setlength{\tabcolsep}{4pt}
\renewcommand{\arraystretch}{0.95}
\begin{tabular}{lll}
\toprule
 & $\epsilon$ off & $\epsilon$ on \\
\midrule
\multicolumn{3}{l}{\textit{Qwen2.5-0.5B / Korean (indep.\ predicts 78.0\%)}} \\
replay 0\% & 0.1373          & 0.0418 (69.5\%) \\
replay 1\% & 0.0993 (27.7\%) & 0.0277 (79.8\%) \\
\midrule
\multicolumn{3}{l}{\textit{OLMo-2-1B / Korean (indep.\ predicts 64.8\%)}} \\
replay 0\% & 0.3997          & 0.2283 (42.9\%) \\
replay 1\% & 0.2460 (38.4\%) & 0.1023 (74.4\%) \\
\bottomrule
\end{tabular}}
\end{minipage}\hfill
\begin{minipage}[t]{0.62\textwidth}
\centering
\scriptsize
\captionof{table}{Best recovery from $\alpha$-sweeping the shared drift ($c(t) = 0$, replay-free; rates and budgets in Table~\ref{tab:geo}). Post-hoc edits recover $<5\%$ of forgetting; this is one family of edit, with row reset and uniform interpolation in Appendix~\ref{app:posthoc}.}
\label{tab:posthoc}
\resizebox{\linewidth}{!}{%
\setlength{\tabcolsep}{4pt}
\renewcommand{\arraystretch}{0.95}
\begin{tabular}{lrrrr}
\toprule
Setting & Rows & Recovery (\%) & Best $\alpha$ & Learn at $\alpha{=}1$ (\%) \\
\midrule
OLMo-2-1B / ko   & 42{,}037 & $+0.7$ & 0.25 & $-133.6$ \\
Qwen-0.5B / ko   & 85{,}143 & $+2.5$ & 0.50 & $-18.7$ \\
Pythia-1.4B / ko & 15{,}542 & $+4.6$ & 0.50 & $-7.7$ \\
\midrule
\multicolumn{5}{l}{\textit{Time course, Qwen / Korean: best recovery by checkpoint (tokens)}} \\
\multicolumn{5}{l}{$+1.7$ (6.7M) \quad $+1.8$ (13.4M) \quad $+2.0$ (20M) \quad $+2.1$ (26.7M) \quad $+2.5$ (33.3M) \quad $+2.5\%$ (40M)} \\
\bottomrule
\end{tabular}}
\end{minipage}
\end{table}

\subsection{Why the Replay-Free Baselines Stop Short}
\label{sec:baselines}

\textbf{Comparison with baselines.}
Without access to $\Dold$, we evaluate five baselines on Qwen/Korean: $L_2$-to-base, KL-to-base, a constant low learning rate, bottom-layer freezing, and a low-rank adapter at $r = 16$; EWC is excluded because estimating Fisher weights requires original data. None reaches the free point: both anchors and the low rate buy retention only by sacrificing target learning, and bottom-layer freezing worsens both axes. They share one cause: all restrict aggregate parameter drift without a coordinate metric separating harmful drift from new learning, the axis that per-parameter $\sqrt{\hat{v}}$ supplies and output-scoped $\epsilon$ exploits. Appendix~\ref{app:baselines} gives the protocol, with per-arm numbers in Table~\ref{tab:base} and the frontier in Figure~\ref{fig:frontier}; the ordering generalizes to OLMo-2-1B (Appendix~\ref{app:base2}).

\textbf{The adapter is the informative failure.}
LoRA tests our premise directly. At its optimal rate it reduces forgetting by only 12.0\% while giving up 27.1\% of target learning, and that retention comes mostly from keeping the output head at base by default: releasing the head at that rate, ten times the full fine-tuning rate, spikes forgetting 23-fold, from 0.121 to 2.820 nats, through one tensor. Output-scoped $\epsilon$ on the released head removes 91.1\% of that at the same rate, while reaching the highest target learning of the adapter arms. Low rank also contributes, though: on TinyLlama, head freezing without an adapter leaves 5.13 nats at the adapter rate (against 0.062 for LoRA) and turns learning negative, while at the full fine-tuning rate head freezing and output-scoped $\epsilon$ converge within 0.001 nats (47.0\% and 47.2\%), so rank constraints protect only at the tenfold higher adapter rates (Appendix~\ref{app:lora}).

\subsection{Acting After Training Fails}
\label{sec:posthoc}

\textbf{The edit does not recover what the dial prevents.}
In our replay-free setting, sweeping $\alpha$ over the shared drift component recovers at most $+4.6\%$ of forgetting across three settings, with the optimal $\alpha$ always representing partial removal; full removal degrades target learning by $-7.7\%$ to $-133.6\%$ (Table~\ref{tab:posthoc}). Editing earlier does not help: across six checkpoints the best recovery moves only from $+1.7\%$ to $+2.5\%$, and shared-direction energy saturates at 66\% by 20M tokens.

\textbf{Why, and it is not an implementation detail.}
Drift and repair share one tensor, and ablating each component separates their roles (Appendix~\ref{app:geo}). Removing the shared direction leaves forgetting essentially unchanged ($0.98\times$ to $1.10\times$) while degrading target learning, so that direction carries new-domain probability mass rather than damage. Ablating the residual instead amplifies forgetting by $1.6\times$, $1.7\times$ and $2.4\times$ while leaving learning within $1.8\%$, so the residual stores the compensation that protects legacy capabilities under a drifting head. Any post-hoc edit strands or erases that compensation; raising $\epsilon$ during training prevents the drift, so no compensation is needed.

\textbf{A second regime, outside our setting.}
Under a 1\% replay regime (outside our data-free scope, Section~\ref{sec:problem}), tracking edit dynamics over time confirms this limitation. Partial row resets recover at most $+3.5\%$ with tied embeddings and nothing with untied ones, while full resets cost rather than recover, monotonically more as training proceeds ($-12.3\%$ to $-68.7\%$ over six checkpoints; Appendix~\ref{app:posthoc}). Neither regime provides an actionable repair window.

\subsection{When the Forgetting Is in the Body}
\label{sec:bodyeps}

When $\Dnew$ introduces no new vocabulary, causal share shifts into the body (Section~\ref{sec:spectrum}). Body-scoped $\epsilon$ on Alpaca at two base models shows the site transfers but the free lunch does not: the body dial removes several times more forgetting than the output dial at both cells, yet is free at only one, paying beyond seed variance at the other. The mechanism explains the asymmetry: an output row for an absent token accumulates pure drift, so damping it costs nothing, whereas a body coordinate encodes new representations and prior capabilities together, so damping it suppresses acquisition. Losslessness is a property of the site, not of $\epsilon$; where sites mix, replay and rate scaling remain the remedies (Appendix~\ref{app:bodyeps}).

\section{Discussion and Limitations}
\label{sec:discussion}

\textbf{Consistency with other remedies.}
Our framework unifies disparate findings along one variable, vocabulary deficit: rare-token gating~\citep{yu2022rare} matches output $\epsilon$, bottom-layer freezing~\citep{zheng2025spurious} fails under output-dominated forgetting, and weight interpolation~\citep{wortsman2022robust} degrades gracefully where selective edits collapse because uniform scaling moves drift and compensation together, though it still trades learning for retention (Appendix~\ref{app:posthoc}).

\textbf{Limitations.}
First, the pre-retraining diagnostic predicts rank order only; absolute forgetting depends on base capacity and resists cross-tokenizer calibration (Appendix~\ref{app:preflight}). Second, in-distribution defenses are preliminary: the inward shift to the body reproduces at both tested cells, but lossless body $\epsilon$ holds at one (Section~\ref{sec:bodyeps}). Third, reduction is not uniform within a setting: at Pythia-410M on Korean two of six seeds recover little and forfeit target learning (Section~\ref{sec:spectrum}), so cost-free mitigation describes most runs, not all. Fourth, scaling comparisons are Pythia-only, the $\epsilon$ sweep covers three Pythia checkpoints (Table~\ref{tab:epsval}), and the capability check predates the grid and runs at $\epsilon = \num{1e-5}$ on math with 1\% replay, not data-free Korean (Appendix~\ref{app:checks}). Fifth, the adapter ablation uses one rank ($r = 16$) at two base models (Appendix~\ref{app:lora}). Sixth, the mechanism needs coordinate-wise adaptive optimizers, leaving Muon~\citep{jordan2024muon} outside its scope.

\textbf{Practical summary.}
Assess vocabulary deficit before retraining; if it is high, raise output $\epsilon$, run a standard rate sweep without tuning $\epsilon$, and add replay when available. No post-hoc edit we tried recovers what prevention achieves.

\section{Conclusion}
\label{sec:conclusion}

We localize catastrophic forgetting under domain shift to the output embeddings of under-exposed tokens, where non-target softmax gradients meet Adam's scale normalization. Raising $\epsilon$ there alone removes 39.4--67.9\% of forgetting across seven stable settings, without original data or lost target learning, and it has to be done during training: drift and compensation share one tensor, so a later selective edit strands or destroys what sustains the old capabilities. The first defense against catastrophic forgetting may lie in optimizer configuration rather than data curation.

\section*{Author Disclosure of Generative AI Tools}
In this work, large language model (LLM) tools were used for figure generation, final checking, translation, and LaTeX formatting. The authors take full responsibility for the content of this paper.

\clearpage
\appendix
\raggedbottom
\renewcommand{\thetable}{A\arabic{table}}
\renewcommand{\theHtable}{appendix.\arabic{table}}
\setcounter{table}{0}
\renewcommand{\thefigure}{A\arabic{figure}}
\renewcommand{\theHfigure}{appendix.\arabic{figure}}
\setcounter{figure}{0}

\section{Appendix}
\label{sec:appendix}
The appendix opens with the tables the main text cites and then groups the deferred details by topic. Section~\ref{app:tables} compiles those supplementary tables, while subsequent sections take the details in this order: method equivalence, the pre-retraining readout, the learning-rate operating regime, capability and length validation, drift geometry and component ablations, the attention column of the instruction grid, the body dial at an in-distribution corpus, low-rank adapters, the replay-free baseline suite and its repetition at a second base model, a second injected domain, the site-by-site second moment, and post-hoc editing.

\subsection{Supporting tables}
\label{app:tables}

\textbf{Learning rates behind the two domain-varying tables.} Both follow the peak-on-learning
rule of Section~\ref{sec:cells}, which selects per cell, so neither table is rate-matched along
a row and neither should be differenced across one. For Table~\ref{tab:domain}: math and code at
\num{3e-5} in all three models except Qwen/math at \num{1e-5}, and Korean at \num{1e-4}
throughout. For Table~\ref{tab:module}, as Alpaca/KoAlpaca: Pythia-410M \num{3e-5} and
\num{3e-5}, Pythia-1.4B \num{1e-5} and \num{1e-4}, Qwen \num{3e-6} and \num{3e-5}. Only the
410M column is therefore rate-matched across the two corpora; the Korean cell of
Table~\ref{tab:domain} also sits at a rate that buys 0.6\% of target learning for roughly four
times the forgetting of \num{3e-5}, which inflates its denominator.

\begin{table}[t]
\caption{Reference forgetting and the composition of the lowest band
($\sqrt{\hat{v}} < \num{1e-6}$) per site: parameters held (M) and, for the input
embedding, the share of that set whose $\sqrt{\hat{v}}$ is exactly zero. Reference forgetting
is this campaign's own pair; at 160M, whose runs scatter widely at \num{3e-4}, the six-seed
mean of Table~\ref{tab:rate160m} is 3.108 nats against the 3.392 here.}
\label{tab:locstats}
\begin{center}\scriptsize
\begin{tabular}{lrrrrrr}
\toprule
Setting & lr & Ref.\ forget & Out.\ params (M) & In.\ params (M) & In.\ zero (\%) & Body params (M) \\
\midrule
160M & \num{3e-4} & 3.392 & 30.7 & 30.9 & 45.8 & 1.2 \\
410M & \num{1e-4} & 0.656 & 38.7 & 41.3 & 45.7 & 2.7 \\
1.4B & \num{1e-4} & 0.472 & 80.6 & 88.0 & 42.8 & 15.9 \\
TinyLlama & \num{1e-4} & 0.355 & 48.2 & 50.4 & 45.3 & 13.1 \\
OLMo & \num{3e-4} & 1.160 & 176.6 & 199.6 & 52.8 & 6.9 \\
Qwen & \num{1e-4} & 0.628 & 110.0 & n/a & n/a & 1.0 \\
\bottomrule
\end{tabular}
\end{center}
\end{table}

\begin{table}[t]
\caption{Pythia-160M across learning rates: output-only $\epsilon$, Korean, replay-free, 40M
tokens. The \num{1e-3} uptick sits inside the seed spread. Cells here are matched pairs drawn
from one campaign, with the same seed count on both arms, whereas the 160M row of
Table~\ref{tab:lrcurve} pools every available reference at each rate into a deeper but
unbalanced denominator; that is why the two differ by up to four points at the shared rates.}
\label{tab:rate160m}
\centering
\scriptsize
\begin{tabular}{lrrrrc}
\toprule
lr & Ref.\ forget (nats) & Interv.\ forget & Reduction (\%) & Learning change (\%) & $n$ (ref/interv) \\
\midrule
\num{1e-5} & 0.491 & 0.126 & 74.2    & $-2.5$  & 3/3 \\
\num{3e-5} & 1.448 & 0.388 & 73.2    & $+9.9$  & 3/3 \\
\num{1e-4} & 2.068 & 1.855 & 10.3    & $-7.5$  & 6/6 \\
\num{3e-4} & 3.108 & 3.795 & $-22.1$ & $-17.3$ & 6/6 \\
\num{1e-3} & 5.233 & 4.681 & 10.5    & $+20.5$ & 2/3 \\
\bottomrule
\end{tabular}
\end{table}

\begin{table}[t]
\caption{Causal share by (site, $\sqrt{\hat{v}}$ band): forgetting reduction (learning change in
parentheses, \%) from freezing that group. Korean, replay-free, 40M tokens, peak lr, two
seeds per arm except Qwen at three. Every column is at that setting's peak rate except $\dagger$, which repeats the 160M
grid at \num{1e-5}: the 160M peak collapses and every band row in its column goes negative
(the two count-matched controls do not), so $\dagger$ is
what Figure~\ref{fig:loc}a plots for that model and what (iv) of Section~\ref{sec:loc} reads.
Each column divides by its own campaign's three-seed reference trained at that column's rate.
Dashes are arms not run. Freezing a band and
freezing by token count are different membership rules, not two estimates of one quantity.
Count-matched rows freeze the output lowest band's mass from the named site's low end.
Stats: Table~\ref{tab:locstats}.}
\label{tab:locfull}
\begin{center}\scriptsize
\resizebox{\linewidth}{!}{%
\begin{tabular}{llccccccc}
\toprule
Site & Band & 160M & 160M$^{\dagger}$ & 410M & 1.4B & TinyLlama & OLMo & Qwen \\
\midrule
Output emb. & $<$\num{1e-6} & -6.2 (-13.4) & 68.8 (+3.1) & 31.4 (+5.2) & 33.7 (+2.0) & 37.0 (+0.1) & 31.3 (+1.6) & 51.0 (+4.4) \\
 & \num{1e-6}--\num{1e-5} & -13.6 (-9.7) & -20.1 (-0.8) & 2.3 (+6.3) & 1.1 (+0.1) & 3.6 (-0.4) & 0.3 (-0.1) & 10.7 (+1.3) \\
 & \num{1e-5}--\num{1e-4} & -10.9 (-7.7) & -17.8 (+2.2) & -5.6 (+0.6) & -1.0 (-0.2) & -3.5 (-0.5) & 1.3 (-0.4) & 1.1 (-0.8) \\
Input emb. & $<$\num{1e-6} & -3.6 (-4.6) & 7.3 (-1.8) & 1.4 (+9.0) & -1.5 (+0.1) & -0.6 (-0.5) & 0.4 (-0.2) & n/a \\
 & \num{1e-6}--\num{1e-5} & -17.9 (-6.5) & 6.8 (-2.4) & 3.3 (+8.1) & -1.5 (+0.2) & -0.3 (-0.4) & 1.9 (-0.2) & n/a \\
 & \num{1e-5}--\num{1e-4} & -18.7 (-9.3) & 12.5 (+0.2) & 7.2 (+9.9) & -0.5 (+0.1) & 1.3 (-0.2) & -1.3 (-0.2) & n/a \\
 & count-matched & 9.8 (+2.3) & --- & 2.6 (+7.2) & -0.5 (+0.0) & 0.4 (-0.3) & -1.0 (-0.2) & n/a \\
Body & $<$\num{1e-6} & -0.1 (-5.1) & 4.6 (+0.4) & 2.7 (+8.6) & -1.3 (+0.3) & 0.2 (-0.5) & 0.7 (-0.3) & 0.2 (+0.0) \\
 & \num{1e-6}--\num{1e-5} & -2.6 (-6.4) & 0.8 (-10.3) & -28.8 (-17.6) & 20.7 (-5.4) & 28.4 (-4.9) & 36.9 (-8.8) & 4.9 (-2.9) \\
 & \num{1e-5}--\num{1e-4} & -5.8 (-39.6) & 33.7 (-7.1) & 56.3 (-40.7) & 43.8 (-23.8) & 57.0 (-13.5) & 25.5 (-1.0) & 26.8 (-36.3) \\
 & count-matched & 28.5 (+4.2) & --- & 11.2 (+7.5) & 0.4 (+0.2) & 2.9 (-0.4) & 4.4 (-0.4) & 12.3 (-9.4) \\
\bottomrule
\end{tabular}}
\end{center}
\end{table}
\begin{table}[t]
\caption{Distribution of $\sqrt{\hat{v}}$ at the end of the reference run (peak lr, Korean,
replay-free, 40M tokens, seed 0). Zero coordinates never received a gradient (unfed input
rows; tied Qwen has none) and carry no update for $\epsilon$ to damp; the two threshold
columns exclude them. The last column shares the zeros of the input table alone
(Table~\ref{tab:locstats}'s In-zero column instead shares them of the input's lowest band).}
\label{tab:v_distribution}
\centering
\scriptsize
\begin{tabular}{lrrrr}
\toprule
Setting & Exactly 0 (\%) & $<\num{1e-8}$ (\%) & $<\num{1e-6}$ (\%) & Zero, input emb.\ (\%) \\
\midrule
Pythia-160M    & 8.7 & 15.4 & 32.8 & 36.6 \\
Pythia-410M    & 4.7 & 4.6  & 16.5 & 36.6 \\
Pythia-1.4B    & 2.7 & 3.5  & 10.7 & 36.6 \\
TinyLlama-1.1B & 2.1 & 3.0  & 8.3  & 34.8 \\
OLMo-2-1B      & 7.1 & 11.2 & 20.1 & 51.2 \\
Qwen2.5-0.5B   & 0.0 & 5.0  & 22.5 & n/a \\
\bottomrule
\end{tabular}
\end{table}

\begin{table}[t]
\caption{Sensitivity to the $\epsilon$ value, Korean injection, replay-free, 40M tokens:
forgetting reduction (learning change in parentheses, both \%) against the same reference.
At 1.4B the response is flat across the hundredfold span; 410M gives up eleven points at the
\num{1e-3} edge; in 160M's collapse regime no value rescues the run. The sweep exists only for
these three Pythia checkpoints, so the flatness claim is not tested on the other three families.}
\label{tab:epsval}
\centering
\scriptsize
\begin{tabular}{llrrr}
\toprule
Setting & lr & $\epsilon = \num{1e-5}$ & $\epsilon = \num{1e-4}$ & $\epsilon = \num{1e-3}$ \\
\midrule
Pythia-160M & \num{3e-4} & $-6.2$ ($-10.7$) & $-22.1$ ($-17.3$) & $-30.0$ ($-24.4$) \\
Pythia-410M & \num{1e-4} & 41.5 ($+13.2$)   & 43.8 ($+10.9$)    & 32.7 ($+2.5$) \\
Pythia-1.4B & \num{1e-4} & 40.9 ($+2.1$)    & 44.8 ($+2.2$)     & 45.6 ($+2.2$) \\
\bottomrule
\end{tabular}
\end{table}

\begin{table}[t]
\centering
\caption{No baseline reaches the free point. Qwen / Korean, replay-free, 40M tokens, three
seeds per arm, all measured against one reference trained at lr \num{3e-5} (forget $+0.1373$
nats, learn $+0.3189$). Arms whose own sweep peaks elsewhere carry their rate in the Setting
column. Qwen's reference learns most at \num{1e-4}; \num{3e-5} is the rate at which this
baseline campaign was run. Qwen ties its input and output embeddings, so on this model
``output-only'' scopes $\epsilon$ to the one shared table and the input/output separation
of Table~\ref{tab:base2} cannot be drawn here.}
\label{tab:base}
\scriptsize
\resizebox{0.8\linewidth}{!}{%
\begin{tabular}{llrr}
\toprule
Method & Setting & Forgetting reduction (\%) & Learning change (\%) \\
\midrule
Output-only $\epsilon$ (ours) & $\epsilon = \num{1e-4}$ & 69.5 & $+0.4$ \\
Under-exposed freeze          & $K = 1{,}000$    & 70.7    & $-1.4$ \\
Step-wise gating              &                  & 70.2    & $-0.4$ \\
$L_2$-to-base                 & $\lambda = 10$   & 11.2    & $-1.3$ \\
$L_2$-to-base                 & $\lambda = 30$   & 29.3    & $-4.4$ \\
$L_2$-to-base                 & $\lambda = 100$  & 64.3    & $-15.4$ \\
KL-to-base                    & $\beta = 0.1$    & 96.5    & $-95.3$ \\
KL-to-base                    & $\beta = 0.3$    & 96.4    & $-97.3$ \\
KL-to-base                    & $\beta = 1$      & 96.4    & $-98.0$ \\
Constant low LR               & \num{3e-6}       & 98.8    & $-53.8$ \\
Bottom-layer freeze           & 6 blocks         & $-2.5$  & $-11.4$ \\
LoRA, $r = 16$                & lr \num{1e-3}    & 12.0    & $-27.1$ \\
\bottomrule
\end{tabular}}

\end{table}

\begin{table}[t]
\caption{Disabling weight decay leaves forgetting and reduction unchanged, confirming drift is not a decay artifact (Korean, replay-free, 40M tokens, lr $10^{-4}$). The decoupled term $\lambda\theta$ of Eq.~\eqref{eq:adam} is therefore inert at this site, which is why Section~\ref{sec:math} reads the adaptive step alone. The 410M gap reverses sign as seeds are added: at three seeds $\mathrm{wd} = 0.1$ led, 39.2\% against 29.0\% for $\mathrm{wd} = 0$, while at six seeds $\mathrm{wd} = 0$ leads, 38.4\% against 32.7\%. A gap that changes sign with three more seeds is noise rather than an effect.}
\label{tab:wd}
\centering
\footnotesize
\setlength{\tabcolsep}{4pt}
\begin{tabular}{@{}llrrc@{}}
\toprule
Base model & wd & Reduct. & Learn. & $n$ \\
\midrule
Pythia-410M  & 0   & 38.4\% & $-5.7$\% & 6 \\
Pythia-410M  & 0.1 & 32.7\% & $-7.0$\% & 6 \\
\addlinespace[2pt]
Qwen2.5-0.5B & 0   & 68.0\% & $+5.7$\% & 3 \\
Qwen2.5-0.5B & 0.1 & 67.9\% & $+5.6$\% & 3 \\
\bottomrule
\end{tabular}
\end{table}

\subsection{Instrument equivalence in detail}
\label{app:equiv}

\begin{figure}[t]
\centering
\includegraphics[width=0.55\linewidth]{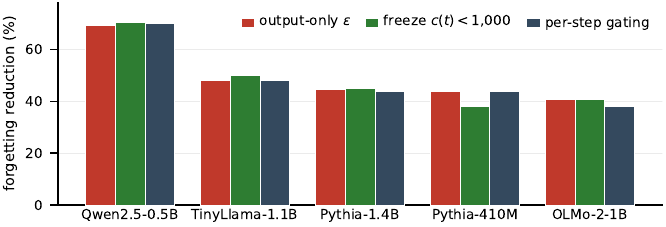}
\caption{Three instruments stopping the same rows achieve equivalent reduction (Korean, replay-free; rates in Table~\ref{tab:equiv}).}
\label{fig:equiv}
\end{figure}

\textbf{Controls beyond the three instruments.}
Table~\ref{tab:equiv} reports the full comparison across models, and Figure~\ref{fig:equiv} plots the three instruments side by side; for Pythia-160M, all instruments yield near-zero or noisy metrics due to optimization collapse. Freezing the same parameter mass under different criteria separates retention from selectivity, and the separation shows up in learning rather than in retention (Table~\ref{tab:mass}). At a matched 96.6\% of the output embedding, in the one campaign that carries 1\% replay, a random mask removes 66.4\% of forgetting and a count mask removes 68.2\%, a gap of under two points; freezing the whole projection removes 74.4\%, so mass is what buys retention. The criterion decides what the mass costs. The random mask pays $-4.4\%$ of target learning, the same price as freezing everything, while the count mask pays $-0.2\%$ and $\epsilon$ pays nothing at all. Which rows stay free, not how many are held, is what makes the retention free. Furthermore, scaling down the embedding learning rate traces the exact trade-off frontier of elevating $\epsilon$: a learning-rate factor of $\times 0.10$ achieves 67.5\% reduction and $\times 0.36$ achieves 51.3\%, with the $\epsilon$ sweep spanning the intermediate range (Table~\ref{tab:lrvseps}). Finally, freezing body parameters across matched $\sqrt{\hat{v}}$ bands yields negligible mitigation until reaching bands that severely degrade target acquisition (41.6\% learning penalty), confirming that the mechanism is strictly confined to under-exposed output embeddings.

\textbf{Random rows against the lowest band.} The contrast quoted in Section~\ref{sec:loc}(ii) comes from one replay-free Qwen campaign at lr \num{3e-5}, three seeds per arm, all divided by the same reference (forget $+0.1374$). Holding 85{,}143 randomly chosen output rows at base removes 12.1\% of forgetting and costs $4.5\%$ of target learning; holding the lowest-$\sqrt{\hat{v}}$ coordinates of the same table removes 42.7\% and gains $1.7\%$. Two cautions attach. The masks are not count-matched: the random arm holds 85{,}143 of 151{,}936 rows, or 76.3M parameters, while the band arm holds 108.8M, so the pair separates selection criteria and not parameter counts, and the count-controlled statement rests on Table~\ref{tab:massmatch} and Table~\ref{tab:mass} instead. The band arm also holds coordinates rather than whole rows, since $\sqrt{\hat{v}}$ membership is per parameter; against the wider ten-seed reference pool of the same cell the same arm reads 43.1\% rather than 42.7\%.

\begin{table}[t]
\caption{Comparison of three equivalence instruments (output-specific $\epsilon = 1 \times 10^{-4}$, freezing rows with $c(t) < 1{,}000$, and step-wise gating) under Korean injection (replay-free, 40M tokens, two seeds per arm except Qwen and every
$\epsilon$ arm at three). The rate column is the campaign rate, which is the peak rate except
at Qwen, whose reference learns most at \num{1e-4}, and at 160M, which has no stable peak.
Cells keep this campaign's seed pool; main-table rows pool more seeds, hence small offsets
(OLMo 40.7 here vs.\ 39.4 in Table~\ref{tab:main}).}
\label{tab:equiv}
\centering
\small
\resizebox{\linewidth}{!}{%
\begin{tabular}{llrrrr}
\toprule
 & & \multicolumn{3}{c}{Forgetting reduction (\%)} & \\
\cmidrule(lr){3-5}
Setting & lr & Output-only $\epsilon$ & Freeze, $K{=}1{,}000$ & Step-wise gating & Max gap (pp) \\
\midrule
Qwen2.5-0.5B    & \num{3e-5} & 69.5 & 70.7   & 70.2   & 1.2 \\
Pythia-1.4B     & \num{1e-4} & 44.8 & 45.1   & 43.7   & 1.5 \\
TinyLlama-1.1B  & \num{1e-4} & 48.1 & 50.1   & 48.2   & 2.0 \\
OLMo-2-1B       & \num{3e-4} & 40.7 & 40.7   & 38.1   & 2.6 \\
Pythia-410M     & \num{1e-4} & 43.8 & 38.1   & 43.7   & 5.7 \\
Pythia-160M     & \num{1e-3} & 8.0  & $-7.6$ & $-4.6$ & 15.6 \\
\bottomrule
\end{tabular}}
\end{table}
\begin{table}[t]
\caption{Retention under different selection criteria at matched mass: every arm below the first freezes 96.6\% of the output embedding, while the first row freezes the whole projection, so only the lower four rows are mass-matched with each other (Qwen/Korean, 40M tokens, lr \num{3e-5}, three seeds, one reference with forget $+0.0993$ nats). This campaign carries 1\% replay, which is why its $\epsilon$ row is the 72.1\% share of Table~\ref{tab:replay} and not the replay-free 69.5\% of Table~\ref{tab:base}.}
\label{tab:mass}
\centering
\small
\begin{tabular}{lrr}
\toprule
Arm & Forgetting reduction (\%) & Learning change (\%) \\
\midrule
Freeze the whole output projection        & 74.4 & $-4.4$ \\
Freeze a random 96.6\% of the embeddings  & 66.4 & $-4.4$ \\
Freeze by count, $K = 100$ (96.6\%)       & 68.2 & $-0.2$ \\
Output-only $\epsilon = \num{1e-4}$       & 72.1 & $+0.3$ \\
Count-freeze and $\epsilon$ together      & 73.2 & $-0.8$ \\
\bottomrule
\end{tabular}
\end{table}
\begin{table}[t]
\caption{Equivalence of embedding learning-rate scaling and $\epsilon$ elevation, alongside body negative controls across $\sqrt{\hat{v}}$ bands (Qwen/Korean, replay-free, three seeds). All arms share one reference, whose learning is 0.3507 and forgetting 0.1924 nats; that reference belongs to this campaign and is not the one behind Table~\ref{tab:base}.}
\label{tab:lrvseps}
\centering
\small
\begin{tabular}{lrr}
\toprule
Arm & Forgetting reduction (\%) & Learning change (\%) \\
\midrule
embedding learning rate $\times 0.10$          & 67.5 & $+0.7$ \\
embedding $\epsilon = \num{5e-5}$              & 66.7 & $+2.2$ \\
embedding $\epsilon = \num{2e-5}$              & 64.2 & $+2.8$ \\
embedding $\epsilon = \num{5e-6}$              & 57.9 & $+3.0$ \\
embedding $\epsilon = \num{2e-6}$              & 52.0 & $+2.8$ \\
embedding learning rate $\times 0.36$          & 51.3 & $+1.4$ \\
\midrule
freeze embedding, $\sqrt{\hat{v}} < \num{1e-6}$        & 45.4 & $+2.4$ \\
freeze embedding, \num{1e-6}--\num{1e-5}               & 16.5 & $+0.8$ \\
freeze embedding, \num{1e-5}--\num{1e-4}               & 1.2  & $-0.9$ \\
freeze body, $\sqrt{\hat{v}} < \num{1e-6}$             & 0.1  & $-0.0$ \\
freeze body, \num{1e-6}--\num{1e-5}                    & 5.4  & $-3.9$ \\
freeze body, \num{1e-5}--\num{1e-4}                    & 19.4 & $-41.6$ \\
\bottomrule
\end{tabular}
\end{table}

\subsection{Reading the risk before retraining}
\label{app:preflight}

\textbf{What the deficiency predicts, and on what evidence.}
Vocabulary deficit is computed entirely without training. Following Section~\ref{sec:problem}, it is the share of old-domain token positions, over the four-domain held-out set minus the injected domain, whose target has fewer than 1{,}000 occurrences in the 40M-token new stream, amounting to 63.2\% for Korean, 55.9\% for code, and 33.9\% for math under the Pythia tokenizer (61.3\%, 52.5\%, and 50.6\% for Qwen). On a shared base model, this fraction accurately predicts the relative order of baseline forgetting. Across a domain-axis grid where model architecture and training recipe are held constant, the predicted rank strictly matches empirical observations for all three models (Table~\ref{tab:domain}), consistent with all four prior (checkpoint, corpus) comparisons.

\textbf{Two registered failures bound the claim.}
First, attempting to predict absolute forgetting magnitudes rather than rank order failed. Across ten configurations spanning diverse base models, the Spearman rank correlation between deficit and forgetting is 0.60, falling short of our pre-registered threshold of 0.7. Because forgetting varies across base architectures under identical data, a data-only diagnostic cannot account for model-level factors. Second, attempting to cheaply compensate for these model factors using base-model loss per token also failed. Discrepant tokenization schemes render per-token cross-entropy incomparable across tokenizers, as exemplified by OLMo. Although OLMo's per-token Korean loss appears close to that of English (2.59 vs.\ 2.46 nats), converting to loss per byte reveals it as the checkpoint least familiar with Korean among the four families evaluated ($1.93\times$ vs.\ $1.35$--$1.74\times$). Even with byte normalization, rank predictive power was not restored. The diagnostic remains strictly a data-side, rank-only tool.

\subsection{The operating regime across learning rates}
\label{app:lrcurve}

\textbf{The plateau and its edges.}
Table~\ref{tab:lrcurve} sweeps output-only $\epsilon$ across learning rates in all eight settings. Forgetting reduction holds steady across the low-rate plateau ($10^{-5}$ to $3 \times 10^{-5}$: 50.7\% and 50.6\% in 410M; 57.6\% and 52.1\% in 1.4B), but collapses at $10^{-3}$, falling to 6.7\% in 1.4B and 8.8\% in 410M; OLMo maintains a wider plateau, retaining 26.9\%. The intermediate $3 \times 10^{-4}$ column is where settings part company rather than where they fail together: 410M is already down to 15.3\% and 12B to 3.4\%, while 1.4B holds 32.5\%, OLMo 34.2\% and Qwen 60.9\%. Target learning mirrors this threshold: 410M changes remain within $\pm 1\%$ across the plateau, but drop by 25.6\% at $3 \times 10^{-4}$ and 49.1\% at $10^{-3}$. The plateau also ensures measurement stability: across the 37 cells with a readable reduction, every cell with a seed spread within 8 percentage points exceeds 27\% reduction, whereas spreads over 17 points yield reductions below 17\%. Two cells break monotonic ordering: Pythia-6.9B at $3 \times 10^{-4}$ (47.0\% vs.\ 41.6\% at $10^{-4}$) reflects optimization breakdown where target learning collapses to $-108\%$, and the Pythia-160M uptick at $10^{-3}$ lies within seed variance (Table~\ref{tab:rate160m}). The same shape recurs at the largest scale: 12B falls from 65.3\% at $3 \times 10^{-5}$ to 3.4\% and 1.6\% at the two highest rates, where the reference itself collapses to 6.0 and 6.6 nats. Both failure modes fall on the same two models: at $10^{-5}$ their references forget too little to divide by, and at the top two rates they break down, so at a 40M budget the window that reads at all is $3 \times 10^{-5}$ to $10^{-4}$, narrower than at 410M or 1.4B where every rate is defined. Outside the plateau both retention and new learning fail, which is why $\epsilon$ cannot replace rate tuning.

\textbf{No per-model $\bm{\epsilon}$ tuning.}
The $\epsilon$ value itself requires no per-model adjustment: reduction is flat across a hundredfold span in 1.4B (40.9\% to 45.6\%) and gives up only 11 points at the $10^{-3}$ edge in 410M (Table~\ref{tab:epsval}).

\begin{table}[t]
\caption{Forgetting reduction and learning change (parentheses, both \%) across learning rates (Korean, replay-free, 40M tokens; 3--10 seeds, independently paired vs.\ Table~\ref{tab:main}). Values $>100\%$ denote old loss below base. \emph{n/q} marks a ratio with no readable meaning, and the reason differs by column: in the reduction position the baseline forgets too little to divide by ($<0.025$ nats), while in the learning position the reference itself ends below its base model on the new domain ($-0.45$ and $-0.63$ nats at 12B), so $\mathrm{change}$ in Eq.~\eqref{eq:reduction} divides by a negative quantity and returns $+13.3\%$ and $+50.9\%$ for arms that in fact learned less.}
\label{tab:lrcurve}
\centering
\small
\setlength{\tabcolsep}{5pt}
\renewcommand{\arraystretch}{0.95}
\begin{tabular}{lrrrrr}
\toprule
Setting & \num{1e-5} & \num{3e-5} & \num{1e-4} & \num{3e-4} & \num{1e-3} \\
\midrule
Pythia-160M    & 73.8 ($-2.9$) & 71.4 ($+6.5$) & 10.0 ($-7.8$) & $-17.7$ ($-16.5$) & 8.0 ($+6.1$) \\
Pythia-410M    & 50.7 ($-1.0$) & 50.6 ($+0.7$) & 43.5 ($+10.3$) & 15.3 ($-25.6$) & 8.8 ($-49.1$) \\
Pythia-1.4B    & 57.6 ($-0.1$) & 52.1 ($+0.6$) & 44.9 ($+2.0$) & 32.5 ($+3.8$) & 6.7 ($+16.0$) \\
TinyLlama-1.1B & n/q ($-0.9$)  & 111.6 ($-0.2$) & 47.7 ($+0.4$) & 27.9 ($+2.3$) & 6.1 ($+11.8$) \\
OLMo-2-1B      & 44.5 ($-3.1$) & 46.0 ($-0.8$) & 41.9 ($+0.5$) & 34.2 ($+1.5$) & 26.9 ($+4.5$) \\
Qwen2.5-0.5B   & 75.4 ($-3.8$) & 69.5 ($+0.4$) & 67.7 ($+5.7$) & 60.9 ($+15.9$) & 30.8 ($-77.5$) \\
Pythia-6.9B    & n/q ($-0.1$)  & 60.2 ($+0.8$) & 41.6 ($+2.4$) & 47.0 ($-108.0$) & 8.5 ($-12.9$) \\
Pythia-12B     & n/q ($+0.1$)  & 65.3 ($+0.9$) & 45.3 ($+3.3$) & 3.4 (n/q) & 1.6 (n/q) \\
\bottomrule
\end{tabular}
\end{table}

\subsection{Capability and length checks}
\label{app:checks}

\textbf{Loss is a conservative measure.}
All measurements thus far are loss-based, so we separately verified whether the lower loss reflects recovered capability rather than mere recalibration. This check predates the grid and differs from the main result in four conditions: $\epsilon = \num{1e-5}$ rather than \num{1e-4}, a math injection rather than Korean, 1\% replay rather than none, and a 60M budget at 1.4B. Its loss-based reduction at 6.9B (73.9\% on its own reference pair) is therefore not the 60.2\% of Table~\ref{tab:main}; what it compares is two measures of one pair of runs, not two settings. On Pythia-6.9B across five tasks, the default baseline loses 3.21 capability points whereas the intervention loses only 0.52, so the capability-based reduction (83.6\%) exceeds the loss-based 73.9\% on the same pair (Table~\ref{tab:bench}). On Pythia-1.4B, where each arm carries three benchmark evaluations, the two measures agree within 0.2 points (49.0\% against 49.1\%). Measuring in terms of loss therefore does not overstate the benefit of the intervention, and it may understate it.

\textbf{The gain does not wash out with extended training.}
The main-result budget is 40M tokens. As a tenfold validation check, we trained Pythia-410M/code up to 400M tokens (replay-free, across three seeds). While the reference baseline forgets 1.2701~nats, output-only $\epsilon$ reduces forgetting by 56.2\% and improves learning by 10.1\%, with seed variance remaining within 2\% along both axes. The advantage is not late-emerging: across all six evaluation checkpoints, the intervention's new-domain loss remains strictly below that of the reference baseline.

\textbf{The same shape on twenty evaluation points.}
A second long run measures the budget axis more finely. On Qwen2.5-0.5B-Instruct with Korean
injected to 400M tokens (lr \num{5e-5}, 1\% replay, $\epsilon = \num{1e-5}$, three seeds per
arm, twenty evaluation points), the reduction rises monotonically at every point, reading
75.1\% at 20M, 82.1\% at 160M and 84.7\% at 400M with seed spreads within two points, while
the intervention's new-domain loss stays at or below the reference from 60M on (learning
change $-1.5\%$ at 20M against $+3.9\%$ at 400M). Its conditions are the capability check's
rather than the grid's, so what this curve carries is the shape, that the dial does not decay
as the budget grows, and not a level comparable with Table~\ref{tab:main}.

\begin{table}[t]
\caption{Capability by benchmark rather than loss, at $\epsilon = \num{1e-5}$ against the default
\num{1e-8}: five likelihood-scored tasks at two Pythia checkpoints. Both campaigns run at lr
\num{1e-4} under 1\% replay on a math injection, 6.9B to 40M tokens and 1.4B to 60M, three
seeded training runs per arm; the 1.4B columns average three benchmark evaluations per arm and
the 6.9B columns one. Accuracy in percent; higher is better.}
\label{tab:bench}
\centering
\small
\begin{tabular}{lrrrrrr}
\toprule
 & \multicolumn{3}{c}{Pythia-6.9B} & \multicolumn{3}{c}{Pythia-1.4B} \\
\cmidrule(lr){2-4}\cmidrule(lr){5-7}
Task & untouched & \num{1e-8} & \num{1e-5} & untouched & \num{1e-8} & \num{1e-5} \\
\midrule
LAMBADA     & 61.05 & 60.00 & 65.17 & 60.78 & 52.53 & 56.79 \\
HellaSwag   & 63.24 & 57.89 & 62.06 & 52.06 & 48.08 & 49.90 \\
PIQA        & 76.22 & 72.09 & 73.23 & 71.00 & 67.19 & 69.04 \\
ARC-easy    & 67.26 & 64.02 & 65.32 & 60.73 & 57.00 & 58.33 \\
WinoGrande  & 61.64 & 59.35 & 61.01 & 56.99 & 55.46 & 56.64 \\
\midrule
average     & 65.88 & 62.67 & 65.36 & 60.31 & 56.05 & 58.14 \\
\bottomrule
\end{tabular}
\end{table}

\subsection{How the drift geometry is computed}
\label{app:geo}

\begin{figure}[t]
\centering
\includegraphics[width=0.7\textwidth]{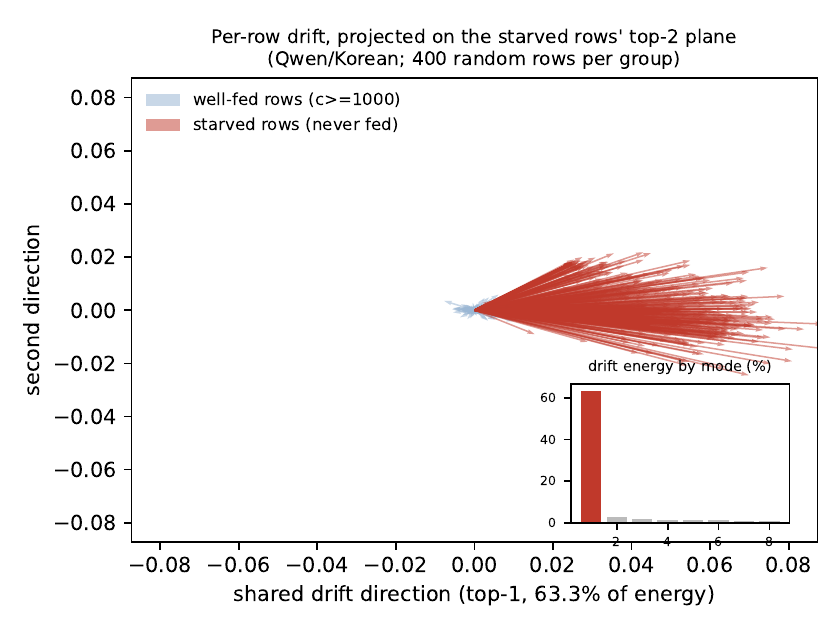}
\caption{Per-token displacement on the top-2 drift plane (Qwen / Korean, 400 rows per group):
absent rows (red) move together, well-fed rows (blue) stay near the origin. Inset:
drift energy by mode.}
\label{fig:current}
\end{figure}

\textbf{The drift is low-rank.}
If the embeddings of absent tokens accumulate a shared push, their aggregate displacement vectors should align along a small number of shared directions rather than scattering independently across embeddings. Computing a singular value decomposition (SVD) on the output projection displacement before and after retraining confirms that displacement energy concentrates predominantly in a single direction (Table~\ref{tab:geo}). Across three Korean-injection settings, the top singular vector of the absent rows accounts for 66.1\% (Qwen), 72.6\% (OLMo), and 80.8\% (Pythia-1.4B) of total energy, compared to merely 3.2--12.9\% in size-matched high-exposure controls ($c(t) \ge 1{,}000$). For embeddings receiving active learning signals, the shared push is diluted by task-specific updates, whereas absent rows retain only the shared push (Figure~\ref{fig:current}).

\begin{table}[t]
\caption{Drift geometry of the output embedding, replay-free checkpoints. Rows: the size of the
absent set ($c(t) = 0$), which is not the number fed to the SVD (Table~\ref{tab:ablate}); that
runs on a size-matched subsample of this set and of the control ($c(t) \ge 1{,}000$). Top-1: energy share on the group's
first singular direction; RMS: mean displacement per row. Rates and budgets, shared with
Tables~\ref{tab:posthoc} and~\ref{tab:ablate}: Qwen \num{3e-5} at 40M, OLMo \num{1e-4} at
60M, Pythia-1.4B/ko \num{1e-4} at 40M, Pythia-1.4B/math \num{3e-5} at 60M.}
\label{tab:geo}
\centering
\scriptsize
\resizebox{\linewidth}{!}{%
\begin{tabular}{lrrrrr}
\toprule
Setting & Rows & Top-1, absent (\%) & Top-1, control (\%) & RMS, absent & RMS, control \\
\midrule
Qwen-0.5B / ko    & 85{,}143 & 66.1 & 3.2  & \num{2.17e-3} & \num{5.53e-4} \\
OLMo-2-1B / ko    & 47{,}492 & 72.6 & 8.8  & \num{5.97e-3} & \num{2.38e-3} \\
Pythia-1.4B / ko  & 15{,}542 & 80.8 & 12.9 & \num{6.30e-3} & \num{1.67e-3} \\
Pythia-1.4B / math & 1{,}242 & 72.0 & 3.7  & \num{3.08e-3} & \num{5.87e-4} \\
\bottomrule
\end{tabular}}
\end{table}

\textbf{Two existing accounts describe the same structure.}
This low-rank structure aligns with two independently proposed accounts. The first stems from the embedding degeneration literature, which observes that rarely occurring token embeddings collapse toward a narrow cone in a single direction~\citep{gao2019representation,yu2022rare}. While those studies analyze models trained from scratch, our measurements capture displacement driven by the same gradient geometry when retraining an already converged base model. The second comes from function-space theory predicting that forgetting concentrates along a small number of modes in output space~\citep{hidekel2026lowrank}. Although we observe equivalent low-rank behavior in parameter-space displacement matrices, whether low-rank structure in parameter space implies that of function space lies beyond our scope.

\textbf{Displacement matrix and singular value decomposition.}
From the pre- and post-retraining output projections $W_{\mathrm{base}}$ and $W_{\mathrm{after}}$, we construct the difference matrix $D = W_{\mathrm{after}} - W_{\mathrm{base}}$ restricted to rows corresponding to target token sets: 85{,}143 rows absent as a target ($c(t) = 0$) for Qwen/Korean and 1{,}242 for Pythia-1.4B/math (Table~\ref{tab:ablate}). From the SVD of $D$, we report the top-1 energy ratio $\sigma_1^2 / \sum_i \sigma_i^2$. The control group applies identical computations to high-exposure embeddings ($c(t) \ge 1{,}000$) within the same model, and both groups are cut to the smaller one's size before the SVD (2{,}080, 794, 381 and 1{,}242 rows by the row order of Table~\ref{tab:geo}).

\textbf{Direction predicted before retraining.}
Running a single forward pass over the new corpus with the base model yields the mean hidden state $\bar{h}$ at the final layer, defining the predicted direction as $-\bar{h}$. A coordinate-scaled variant accounts for Adam by dividing each coordinate by its mean $\sqrt{\hat{v}}$ before unit normalization. Directional alignment is measured by the absolute cosine similarity $|\cos|$ with the empirical top singular vector; the random expectation in dimension $d$ is approximately $1/\sqrt{d}$.

\textbf{Component ablation.}
We decompose each embedding's displacement into its projection along the top singular vector and an orthogonal residual, subtract one component from the retrained weights, and re-evaluate on the validation corpus. Ablating the shared component projects out the top-1 direction from $D$, whereas ablating the residual retains only the top-1 direction. Both variants compare post-ablation target learning and forgetting against the unablated retrained model.

\textbf{How well the predicted direction matches.}
The cosine between the pre-training prediction $-\bar{h}$ and the empirical top singular
vector reaches 0.40 in Qwen/Korean (0.50 under coordinate scaling) and 0.647 in
Pythia-1.4B/math, substantially exceeding the random expectation $1/\sqrt{d}$ (0.033 at
$d = 896$ for Qwen, 0.022 at $d = 2048$ for 1.4B). Neither setting reaches our
pre-registered threshold of 0.7: the consistent negative sign corroborates the
mechanism, but the prediction lacks the precision of an operational tool.

\textbf{The two components do different jobs, but the split is setting-dependent.}
Residual ablation behaves consistently across all three settings: target learning remains largely intact ($-0.9\%$ to $-1.8\%$), whereas forgetting escalates by factors of $1.6\times$ (Pythia-1.4B), $1.7\times$ (OLMo), and $2.4\times$ (Qwen). In contrast, shared component ablation leaves forgetting essentially unaffected ($0.98\times$ to $1.10\times$) while degrading target domain learning by widely varying degrees, spanning $-7.7\%$ at Pythia-1.4B to $-133.6\%$ at OLMo (Table~\ref{tab:ablate}). Although the shared drift consistently contributes to boosting new-domain token probabilities across all three settings, the magnitude of this share varies; hence, our main text claims only the protective function of the residual. In 1.4B/math where vocabulary deficit is minimal, absent rows number only 1{,}242, rendering aggregate metrics insensitive to ablating either component: an exception reflecting measurement limits rather than a contrary mechanism.

\begin{table}[t]
\caption{Component ablation: the shared (top-1) or residual component is subtracted from the
trained weights and the model re-evaluated; negative recovery means forgetting grows. Rows and
top-1 differ from Table~\ref{tab:geo} because that table subsamples both groups to the smaller
one (794 rows at OLMo) while the ablation acts on every absent row, and because the OLMo row
counts $c(t)$ over a 40M window there against the run's full 60M here.}
\label{tab:ablate}
\centering
\scriptsize
\begin{tabular}{lrrrrrr}
\toprule
 & & & \multicolumn{2}{c}{Shared removed} & \multicolumn{2}{c}{Residual removed} \\
\cmidrule(lr){4-5}\cmidrule(lr){6-7}
Setting & Rows & Top-1 (\%) & Recovery (\%) & Learning (\%) & Recovery (\%) & Learning (\%) \\
\midrule
Qwen-0.5B / ko    & 85{,}143 & 66.1 & $-1.7$ & $-18.7$  & $-138.9$ & $-1.8$ \\
OLMo-2-1B / ko    & 42{,}037 & 74.6 & $-9.6$ & $-133.6$ & $-74.4$  & $-1.7$ \\
Pythia-1.4B / ko  & 15{,}542 & 80.3 & $+1.6$ & $-7.7$   & $-60.4$  & $-0.9$ \\
Pythia-1.4B / math & 1{,}242 & 72.0 & $-0.0$ & $-0.1$   & $-1.2$   & $-0.0$ \\
\bottomrule
\end{tabular}
\end{table}

\begin{table}[t]
\caption{The reduction does not shrink with budget: evaluation checkpoints taken inside 160M
token runs, Korean, replay-free; three seeded runs per arm at 6.9B at its peak rate
\num{3e-5}, one at 12B at \num{1e-5}.
Learning changes at the same checkpoints stay within $+2.2\%$ (6.9B) and $+0.6\%$ (12B). The
40M row is a checkpoint of a run still scheduled to 160M, so its learning rate has not yet
decayed; this is why it forgets more than the independent 40M runs of Table~\ref{tab:main} and
why neither those nor the matching rate column of Table~\ref{tab:lrcurve} are comparable with it.}
\label{tab:budget}
\centering
\small
\begin{tabular}{lrrrrrr}
\toprule
 & \multicolumn{3}{c}{Pythia-6.9B} & \multicolumn{3}{c}{Pythia-12B} \\
\cmidrule(lr){2-4}\cmidrule(lr){5-7}
Budget & Ref.\ forget & Interv. & Reduction & Ref.\ forget & Interv. & Reduction \\
\midrule
40M  & 0.2383 & 0.1100 & 53.9\% & 0.0443 & 0.0105 & 76.2\% \\
80M  & 0.3106 & 0.1283 & 58.7\% & 0.0703 & 0.0149 & 78.9\% \\
120M & 0.3114 & 0.1192 & 61.7\% & 0.0757 & 0.0125 & 83.5\% \\
160M & 0.3051 & 0.1152 & 62.3\% & 0.0755 & 0.0120 & 84.1\% \\
\bottomrule
\end{tabular}
\end{table}

\subsection{On the attention column of the instruction grid}
\label{app:module}

One column does not fit the summary above and we state it rather than leave it to be found.
Freezing attention on KoAlpaca removes 42.7\% of forgetting at 410M and 30.2\% at 1.4B, at no
measurable learning cost. At 410M that exceeds what output-scoped $\epsilon$ removes (29.8\%);
at 1.4B it does not (33.1\%), so the two are not separated by a consistent ordering either.
Two things follow. First, causal share does not partition: freezing any large group also stops
whatever that group would have done downstream to the output rows, so shares measured this way
sum past one hundred and a large attention share is not evidence against an output account. The
count-matched controls of Section~\ref{sec:loc} exist precisely because unmatched module
freezing cannot separate site from mass, and attention here is neither count-matched nor
row-resolved. Second, the claim this paper defends about instruction tuning is narrower than
``the output dominates''. It is that the share which is recoverable by a row-level rule moves
to the output when the corpus brings new vocabulary, which is what the $\epsilon$ row shows
(single digits on Alpaca against 29.8--43.3\% on KoAlpaca) and what the module rows cannot
show. Whether attention holds an independent share of instruction-tuning forgetting is a
question our design does not answer, and we do not claim it does.

The output column of the same table carries a second contrast worth stating. At 410M on
KoAlpaca, freezing the whole output module reads $-3.7\%$ while scoping $\epsilon$ to its
low-$\sqrt{\hat{v}}$ rows reads $+29.8\%$, though the freeze is the harder instrument on every
row the two share. What separates them is the rows they do not share: a module freeze also
pins the fed rows, and Appendix~\ref{app:geo} finds those rows carrying the compensation that
holds the old domain up under a drifting head, so stopping the drift and the compensation
together nets to nothing. Blocked acquisition is not the explanation, since the module freeze
leaves target learning unharmed at both untied models ($+3.0\%$ at 410M and $+0.5\%$ at 1.4B);
what it removes is a repair, not a gain.

\subsection{In-distribution forgetting and the body dial}
\label{app:bodyeps}

Section~\ref{sec:spectrum} shows that an in-distribution corpus leaves the causal share in the
body rather than the output. That raises the obvious follow-up: if the site moves, does the
same dial work at the new site? Table~\ref{tab:bodyeps} answers it in two cells, both Alpaca
under instruction fine-tuning at 12.6M tokens and lr \num{1e-5}, replay-free, three seeds per
arm.

In both settings, the body dial eliminates roughly three to seven times more forgetting than the output dial, confirming that in-distribution forgetting resides in the body. However, while the body dial is cost-free within seed noise for Qwen ($-0.3\%$ target learning change for 35.6\% reduction), it imposes a $-6.7\%$ learning penalty on Pythia-1.4B, exceeding measured seed variance. Thus, the localization holds across the two models and losslessness in only one, which is how Section~\ref{sec:bodyeps} reads it.

\begin{table}[t]
\caption{Scoped $\epsilon$ on in-distribution Alpaca across sites (replay-free, 12.6M tokens, lr \num{1e-5}, 3 seeds). Each cell is measured against its own reference: Qwen forget $+0.0620$ nats and learn $+0.1685$; Pythia-1.4B forget $+0.1356$ and learn $+0.3256$.}
\label{tab:bodyeps}
\centering
\small
\begin{tabular}{llrr}
\toprule
Setting & Arm & Forgetting reduction (\%) & Learning change (\%) \\
\midrule
Qwen2.5-0.5B / Alpaca & output $\epsilon = \num{1e-4}$ & 12.1 & $+0.8$ \\
                      & body $\epsilon = \num{1e-5}$   & 14.5 & $+3.0$ \\
                      & body $\epsilon = \num{1e-4}$   & 35.6 & $-0.3$ \\
\midrule
Pythia-1.4B / Alpaca  & output $\epsilon = \num{1e-4}$ & 5.0  & $-0.3$ \\
                      & body $\epsilon = \num{1e-4}$   & 32.5 & $-6.7$ \\
\bottomrule
\end{tabular}
\end{table}

Three details belong here rather than in the main text. First, each cell is paired inside its
own campaign: the Qwen reference is the one its launch script names, and its output-scoped arm
reproduces the 12.1\% that script predicted before the runs were made. Second, the body dial is
dose-dependent in the direction the denominator implies: at Qwen, $\epsilon = \num{1e-5}$
removes 14.5\% against 35.6\% at \num{1e-4}, so the body's larger $\sqrt{\hat{v}}$ needs a
larger floor before it brakes at all, which is the same ordering Table~\ref{tab:vsite} shows
between the sites. Third, we ran the body dial at 1.4B at two higher rates as well, but those
arms have no reference trained at their own rate, so we do not convert them into reductions and
do not report them; the same rule excludes several other unpaired arms from this paper.
Section~\ref{sec:discussion} registers body-scoped $\epsilon$ as preliminary on the strength of
the one lossless cell.

\subsection{Low-rank adapters, and where their retention comes from}
\label{app:lora}

\textbf{The adapter test.}
A low-rank adapter is the closest data-free baseline to our claim, because it constrains
\emph{how much} the weights may move without saying \emph{which} rows may move. If forgetting
lives where we say it lives, then an adapter's retention should come from the output head being
left alone rather than from the low rank, and it should evaporate the moment the head is
allowed to move. Table~\ref{tab:lora} runs that test at Qwen2.5-0.5B on Korean, replay-free, at
40M tokens. The adapter is swept over its own rates and peaks at $10^{-3}$ on target learning
(0.1224, 0.1822, \textbf{0.2329}, 0.2262 at $10^{-4}$ through $3 \times 10^{-3}$); all three
adapter arms share that rate, so they are comparable with each other.

\textbf{What it shows.}
With the head held at base, the default configuration for an adapter, forgetting is 0.1210
nats. Releasing the head to the same adapter returns 2.8204 nats, twenty-three times as much,
from a change that touches one tensor, and output-scoped $\epsilon$ at that released head
removes 91.1\% of that forgetting while leaving the highest target learning of the three
arms.

Three caveats bound how far this reads. First, it is a within-adapter contrast. The
single variable between the two arms is whether the head may move, so the head is what causes
the twenty-three-fold gap; and a matched-rate run that freezes the head
\emph{without} an adapter now separates the two terms: at TinyLlama it leaves 5.1293 nats
against that model's adapter arm at 0.0623 and a negative target learning, so the rank
constraint carries a share of the retention at this rate. The head carries the gap between the two adapter arms, not the whole of the adapter's retention. Second, forgetting at the released head exceeds even the full fine-tuning
reference, because the adapter runs at a rate ten times higher; only the three adapter arms may
be compared with one another. Third, Qwen ties its embeddings, so releasing the head makes a
separate trainable copy of the shared table and breaks the tie, which means the released arm
changes two things at once. TinyLlama unties them and reproduces the contrast: releasing the head there multiplies forgetting 16.3-fold, and output-scoped $\epsilon$ removes 92.4\% of that at the highest target learning of the three arms. The effect is therefore not an artefact of tying.

\begin{table}[t]
\caption{Low-rank adapters at two base models on Korean, replay-free, 40M tokens, $r = 16$,
adapter arms at their swept peak lr \num{1e-3}, three seeds per arm; the head-freeze controls
carry no adapter. Qwen ties its embeddings
and TinyLlama does not. The Qwen full fine-tuning rows are at lr \num{1e-4} and are shown for
scale only. Forgetting and learning in nats.}
\label{tab:lora}
\centering
\small
\begin{tabular}{lrrc}
\toprule
Arm & Forgetting & Learning & $n$ \\
\midrule
\multicolumn{4}{l}{\textit{Qwen2.5-0.5B, tied}} \\
Full fine-tuning reference                & $+0.6285$ & $+0.3764$ & 3 \\
Full fine-tuning $+$ output $\epsilon$    & $+0.2019$ & $+0.3968$ & 3 \\
\addlinespace[2pt]
Adapter, output head frozen               & $+0.1210$ & $+0.2326$ & 3 \\
Adapter, output head released             & $+2.8204$ & $+0.1938$ & 3 \\
Adapter, head released $+$ output $\epsilon$ & $+0.2517$ & $+0.2660$ & 3 \\
\addlinespace[4pt]
\multicolumn{4}{l}{\textit{TinyLlama-1.1B, untied}} \\
Adapter, output head frozen               & $+0.0623$ & $+0.3213$ & 3 \\
Adapter, output head released             & $+1.0179$ & $+0.3186$ & 3 \\
Adapter, head released $+$ output $\epsilon$ & $+0.0773$ & $+0.3356$ & 3 \\
\addlinespace[2pt]
Head frozen, no adapter                   & $+5.1293$ & $-0.1371$ & 3 \\
\addlinespace[4pt]
\multicolumn{4}{l}{\textit{TinyLlama-1.1B, untied, at full fine-tuning's own lr \num{1e-4}}} \\
Full fine-tuning reference                & $+0.3550$ & $+0.4456$ & 3 \\
Head frozen, no adapter                   & $+0.1883$ & $+0.4384$ & 3 \\
Full fine-tuning $+$ output $\epsilon$    & $+0.1874$ & $+0.4474$ & 3 \\
\bottomrule
\end{tabular}
\end{table}

\subsection{The replay-free baseline suite in detail}
\label{app:baselines}

\begin{figure}[t]
\centering
\includegraphics[width=0.82\textwidth]{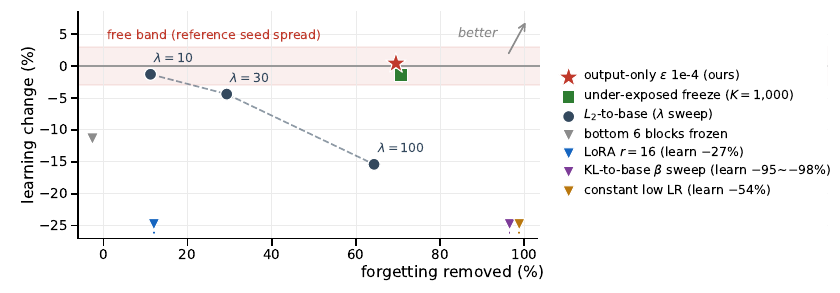}
\caption{Reduction--learning plane (Qwen/Korean, replay-free, 3 seeds). Shaded band is reference seed spread; methods below the axis are marked at bottom and quantified in legend.}
\label{fig:frontier}
\end{figure}

\textbf{Protocol.}
The five baselines of Section~\ref{sec:baselines} are evaluated on Qwen/Korean (vocabulary deficit 61.3\%): $L_2$-to-base (weight penalty $\lambda$ caching base weights), KL-to-base (divergence penalty $\beta$ via parallel base forward passes), a constant low learning rate without reheating, freezing the lowest six blocks, and a low-rank adapter at $r = 16$. EWC is excluded because estimating Fisher weights requires original data that this setting defines as unavailable. Across three seeds at a 40M token budget, each arm trains at its sweep-selected peak rate against a single tuned reference (Figure~\ref{fig:frontier}; Table~\ref{tab:base}).

\textbf{Four ways to stop short.}
$L_2$-to-base presents a rigid trade-off, sacrificing 15.4\% of target learning for 64.3\% forgetting reduction because it restrains under-exposed embeddings and task-active weights indiscriminately. KL-to-base arrests adaptation entirely, eliminating 96.4\% to 96.5\% of forgetting but discarding 95.3\% to 98.0\% of learning, since anchoring to base predictions directly contradicts learning new domain representations. A constant low rate similarly halts adaptation by under-training, forfeiting 53.8\% of target learning. Bottom-layer freezing fails on both fronts, worsening forgetting by 2.5\% while losing 11.4\% of learning.

\subsection{The baseline suite at a second base model}
\label{app:base2}

Table~\ref{tab:base} compares baselines at one setting. Table~\ref{tab:base2} repeats the suite
at OLMo-2-1B, the setting with the largest reference forgetting in the grid. Two conditions differ. The OLMo baseline campaign was run at 1\% replay, the only regime in
which every arm exists, so its reductions are measured against a reference that already carries
a little replay; and it ran to 60M tokens rather than 40M. Neither difference is comparable
cell by cell with Table~\ref{tab:base}. The ordering is what transfers. Output-scoped $\epsilon$, under-exposed
freezing and step-wise gating again land together and again cost nothing; the $L_2$ anchor
reaches two thirds of that reduction only at $\lambda = 10$ and pays for it; the functional
anchor and the constant low rate again trade essentially all of the target domain away; and
bottom-layer freezing is again both weak and expensive. Input-scoped $\epsilon$, the negative
control, does nothing, which is the same null this paper reports from freezing.

\begin{table}[t]
\caption{The baseline suite at OLMo-2-1B / Korean, lr \num{1e-4}, 60M tokens, 1\% replay, three
seeds per arm, against one reference (forget $+0.2455$ nats, learn $+0.7909$). Reductions above 100\% mean the arm left the old-domain
loss below the base model's.}
\label{tab:base2}
\centering
\scriptsize
\resizebox{0.8\linewidth}{!}{%
\begin{tabular}{llrr}
\toprule
Method & Setting & Forgetting reduction (\%) & Learning change (\%) \\
\midrule
Output-only $\epsilon$ (ours) & $\epsilon = \num{1e-4}$ & 58.3   & $+0.7$ \\
Under-exposed freeze          & $K = 1{,}000$   & 58.4   & $+0.6$ \\
Step-wise gating              &                 & 57.6   & $+0.7$ \\
Input-only $\epsilon$         & $\epsilon = \num{1e-4}$ & $-0.5$ & $-0.0$ \\
$L_2$-to-base                 & $\lambda = 0.1$ & 0.5    & $-0.1$ \\
$L_2$-to-base                 & $\lambda = 1$   & 5.8    & $-0.3$ \\
$L_2$-to-base                 & $\lambda = 10$  & 42.0   & $-3.1$ \\
KL-to-base                    & $\beta = 0.1$   & 87.3   & $-95.3$ \\
KL-to-base                    & $\beta = 1$     & 86.2   & $-97.7$ \\
Constant low LR               & \num{3e-6}      & 100.4  & $-52.9$ \\
Bottom-layer freeze           & 6 blocks        & 11.3   & $-10.8$ \\
\bottomrule
\end{tabular}}
\end{table}

\subsection{The same intervention on a second injected domain}
\label{app:code}

Korean is the main injection axis of the grid, chosen so that the base model, the budget and
the evaluation set stay fixed while only the corpus changes. Because Korean sits at the
vocabulary-deficient extreme, Table~\ref{tab:code} collects every code injection we ran, the
next most deficient domain (55.9\% against 63.2\% under the Pythia tokenizer). The reduction
holds across four base checkpoints, across a tenfold range of budget and at two learning rates
on the same checkpoint, and target learning is never meaningfully lost. Two rows deserve their
own reading. At 160M the intervention drives old-domain loss below the base model's, which is
why its reduction exceeds 100\%. And 410M at 400M tokens is the tenfold budget check of
Appendix~\ref{app:checks}: the reduction there is larger than at 40M on the same checkpoint and
the same rate, not smaller.

\begin{table}[t]
\caption{Code injection, replay-free, output-only $\epsilon = \num{1e-4}$, arms paired within
campaign. Rows above the rule are at each campaign's own rate. Above 100\% the old-domain loss
ended below the base model's.}
\label{tab:code}
\centering
\scriptsize
\begin{tabular}{llrrrrc}
\toprule
Setting & lr & Budget & Ref.\ forget & Interv.\ forget & Reduction (\%) & Learning change (\%) \\
\midrule
Pythia-160M  & \num{1e-5} & 40M  & $+0.1876$ & $-0.0287$ & 115.3 & $-0.6$ \\
Pythia-410M  & \num{3e-5} & 40M  & $+0.0803$ & $+0.0356$ & 55.7  & $+0.4$ \\
Pythia-410M  & \num{1e-4} & 40M  & $+0.3248$ & $+0.1938$ & 40.3  & $-0.9$ \\
Pythia-1.4B  & \num{3e-5} & 40M  & $+0.0832$ & $+0.0391$ & 53.1  & $+4.5$ \\
Qwen2.5-0.5B & \num{3e-5} & 40M  & $+0.1806$ & $+0.0493$ & 72.7  & $+18.4$ \\
\midrule
Pythia-410M  & \num{1e-4} & 400M & $+1.2701$ & $+0.5557$ & 56.2  & $+10.1$ \\
\bottomrule
\end{tabular}
\end{table}

\subsection{Where the second moment sits, site by site}
\label{app:vsite}
\begin{table}[t]
\caption{Mean path length per parameter over 120 steps, whole-model $\sqrt{\hat{v}}$ bands,
replay-free, at the campaign rate (410M and 1.4B \num{1e-4}, Qwen \num{3e-5}; Qwen's own peak
is \num{1e-4}). Damping is the ratio of the two distances; net/total is
at the default $\epsilon$ (1.0 = pushed the same way every step, $0.09$ = random walk).}
\label{tab:travel}
\centering
\scriptsize
\resizebox{0.82\linewidth}{!}{%
\begin{tabular}{llrrrrr}
\toprule
Setting & $\sqrt{\hat{v}}$ range & Share & Dist.\ ($\epsilon{=}\num{1e-8}$)
        & Dist.\ ($\epsilon{=}\num{1e-4}$) & Damping & Net/total \\
\midrule
Pythia-410M / ko  & $<\num{1e-6}$            & 6.0\%  & \num{1.75e-3} & \num{3.59e-5} & 48.7 & 0.702 \\
                  & \num{1e-6}--\num{1e-5}   & 14.5\% & \num{2.25e-3} & \num{9.95e-5} & 22.6 & 0.453 \\
                  & \num{1e-5}--\num{1e-4}   & 77.5\% & \num{2.49e-3} & \num{5.60e-4} & 4.4  & 0.320 \\
                  & $>\num{1e-4}$            & 2.0\%  & \num{2.43e-3} & \num{1.38e-3} & 1.8  & 0.288 \\
\midrule
Pythia-1.4B / ko  & $<\num{1e-6}$            & 3.3\%  & \num{1.57e-3} & \num{3.27e-5} & 48.0 & 0.673 \\
                  & \num{1e-6}--\num{1e-5}   & 28.9\% & \num{2.36e-3} & \num{1.55e-4} & 15.2 & 0.353 \\
                  & \num{1e-5}--\num{1e-4}   & 67.5\% & \num{2.47e-3} & \num{3.43e-4} & 7.2  & 0.313 \\
                  & $>\num{1e-4}$            & 0.3\%  & \num{2.34e-3} & \num{1.43e-3} & 1.6  & 0.249 \\
\midrule
Qwen2.5-0.5B / ko & $<\num{1e-6}$            & 21.3\% & \num{1.64e-3} & \num{8.67e-6} & 189.2 & 0.895 \\
                  & \num{1e-6}--\num{1e-5}   & 50.4\% & \num{8.56e-4} & \num{4.22e-5} & 20.3 & 0.456 \\
                  & \num{1e-5}--\num{1e-4}   & 27.7\% & \num{7.81e-4} & \num{1.34e-4} & 5.8  & 0.419 \\
                  & $>\num{1e-4}$            & 0.6\%  & \num{7.64e-4} & \num{5.32e-4} & 1.4  & 0.433 \\
\bottomrule
\end{tabular}}
\end{table}

Table~\ref{tab:travel} bands the whole model at once, so it shows that Adam erases gradient
scale but cannot by itself show that the erasure is an output phenomenon.
Table~\ref{tab:vsite} splits the same quantity by site at the end of the reference run. The two
distributions barely overlap. 74 to 86 percent of the output embedding sits below
\num{1e-6}, against 0.3 to 1.4 percent of the body, and the median coordinate
differs by two to three orders of magnitude between the two: \num{1.1e-8} to \num{7.8e-8} in
the output against \num{5.5e-6} to \num{4.2e-5} in the body. A single scalar placed at
\num{1e-4} therefore leaves an output coordinate under a thousandth of its step while a body
coordinate keeps a twentieth to a third of its own, and that hundredfold gap, not a threshold
either site sits under, is what makes one number selective without a token list. The causal counterpart
of this table is Table~\ref{tab:locfull}, which freezes the same sites in the same bands, and
Table~\ref{tab:lrvseps}, which does so within one setting. What the present data cannot do is
split the travel distances of Table~\ref{tab:travel} the same way, because the probe that
produces them flattens across tensors; that measurement is left open.

\begin{table}[t]
\caption{Distribution of $\sqrt{\hat{v}}$ within each site at the end of the reference run
(Korean, replay-free, 40M tokens, seed 0, campaign rate, which is \num{3e-5} at Qwen against its
peak \num{1e-4}, so its band shares do not reproduce the peak-rate mass held in
Table~\ref{tab:massmatch}). Band columns are percentages of that
site, computed from the float32 dump of the final second moment; the Zero column counts
coordinates that are exactly zero and is contained in the $<\num{1e-6}$ column beside it, so
the two must not be added. Qwen ties its embeddings and has a single table. Medians are over
nonzero coordinates.}
\label{tab:vsite}
\centering
\scriptsize
\setlength{\tabcolsep}{3pt}
\begin{tabular}{llrrrrrr}
\toprule
Setting & Site & Params (M) & Zero & $<\num{1e-6}$ & \num{1e-6}--\num{1e-5} & \num{1e-5}--\num{1e-4} & Median \\
\midrule
Pythia-160M    & output emb. & 38.6  & \textbf{0.0} & 79.4 & 16.3 & 3.9  & \num{2.6e-8} \\
               & input emb.  & 38.6  & 36.6 & 79.9 & 14.9 & 4.7  & \num{1.1e-7} \\
               & body        & 85.1  & 0.0  & 1.4  & 8.1  & 76.5 & \num{4.2e-5} \\
\addlinespace[2pt]
Pythia-410M    & output emb. & 51.5  & \textbf{0.0} & 75.2 & 18.2 & 6.2  & \num{6.8e-8} \\
               & input emb.  & 51.5  & 36.6 & 80.1 & 15.8 & 3.8  & \num{1.0e-7} \\
               & body        & 302.3 & 0.0  & 0.9  & 8.6  & 89.7 & \num{2.6e-5} \\
\addlinespace[2pt]
Pythia-1.4B    & output emb. & 103.0 & \textbf{0.0} & 78.2 & 17.7 & 3.7  & \num{4.5e-8} \\
               & input emb.  & 103.0 & 36.6 & 85.4 & 13.3 & 1.1  & \num{4.9e-8} \\
               & body        & 1208.6& 0.0  & 1.3  & 56.8 & 41.8 & \num{9.1e-6} \\
\addlinespace[2pt]
TinyLlama-1.1B & output emb. & 65.5  & \textbf{0.0} & 73.5 & 22.1 & 3.9  & \num{7.8e-8} \\
               & input emb.  & 65.5  & 34.8 & 76.9 & 18.4 & 4.2  & \num{2.4e-7} \\
               & body        & 969.0 & 0.0  & 1.4  & 33.1 & 65.4 & \num{1.1e-5} \\
\addlinespace[2pt]
OLMo-2-1B      & output emb. & 205.5 & \textbf{0.0} & 85.9 & 11.6 & 2.1  & \num{1.1e-8} \\
               & input emb.  & 205.5 & 51.2 & 97.1 & 2.7  & 0.2  & \num{5.1e-9} \\
               & body        & 1073.9& 0.0  & 0.6  & 90.7 & 8.7  & \num{5.5e-6} \\
\addlinespace[2pt]
Qwen2.5-0.5B   & embedding   & 136.1 & \textbf{0.0} & 82.3 & 11.6 & 5.1  & \num{2.0e-8} \\
               & body        & 357.9 & 0.0  & 0.3  & 15.1 & 83.6 & \num{1.8e-5} \\
\bottomrule
\end{tabular}
\end{table}

\subsection{Post-hoc edits in detail}
\label{app:posthoc}

\begin{figure}[t]
\centering
\includegraphics[width=0.62\textwidth]{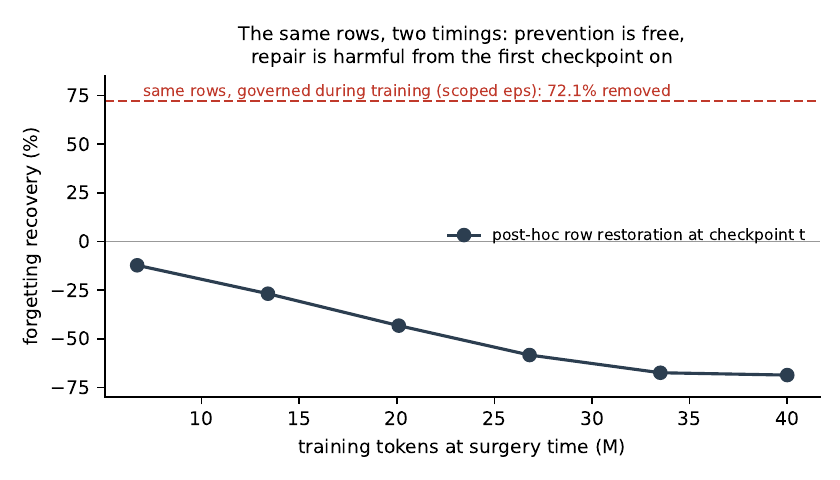}
\caption{The same embeddings at two moments, Qwen / Korean under 1\% replay. Governed during
training, 72.1\% of the forgetting is removed at no cost (dashed, the $\epsilon$ arm of
Table~\ref{tab:mass}); restored after training, the edit is a loss from the first
checkpoint on and degrades monotonically thereafter.}
\label{fig:timing}
\end{figure}

\textbf{Restoration, and its timing.}
We evaluate post-hoc editing under two formulations. First, subtracting the shared component of absent-row drift by ratio $\alpha$ yields maximum recoveries of merely $+0.7\%$, $+2.5\%$, and $+4.6\%$ across three replay-free settings; in all cases, the optimal setting corresponds to partial removal ($\alpha \in [0.25, 0.5]$), whereas full removal ($\alpha = 1$) degrades target domain learning by $-7.7\%$ to $-133.6\%$ (Table~\ref{tab:posthoc}). Second, resetting entire embedding rows to base values achieves at most $+3.5\%$ recovery under 1\% replay (with tied embeddings, $\alpha \approx 0.2$) and 0\% for untied embeddings, with complete reset strictly degrading performance. Temporal dynamics vary across replay regimes. Under zero replay, maximum recovery remains flat between $+1.7\%$ and $+2.5\%$ across all six checkpoints, while energy along the shared direction saturates at 66\% within just 20M tokens. Under 1\% replay, a different arm answers the timing question: restoring the never-fed rows in full ($\alpha = 1$) at each of six saved points leaves recovery negative throughout and monotonically worse, from $-12.3\%$ at 6.7M tokens to $-68.7\%$ at 40M with no inversion (Figure~\ref{fig:timing}). That trajectory is the complete reset, not the partial best-$\alpha$ arm above, which is why its numbers are negative where the best-$\alpha$ recoveries are positive. Neither regime exhibits an early window for viable repair.

\textbf{Why it fails, and the ordering of the damage.}
Component ablation in Appendix~\ref{app:geo} exposes the root cause. Because ablating the residual amplifies forgetting by factors of $1.6\times$, $1.7\times$, and $2.4\times$ across the three settings, compensatory adjustments that sustain legacy capabilities reside within the residual of embedding displacement. We also asked whether the body, rather than the embedding residual, is the main site of that compensation, and the answer is that we could not confirm it either way. The alignment between the shift in legacy-domain mean hidden states and the shared drift axis is $\cos = +0.248$ under 1\% replay, eight times the random expectation of 0.03 but below the 0.3 we registered in advance. We therefore do not claim the body is uninvolved; we claim only that the embedding residual is where the positive evidence is, since ablating it is what multiplies forgetting while ablating the shared direction does not.

The account makes a prediction that can fail: post-hoc damage should scale with how selectively an edit breaks the relationship between drift and compensation, not with how much weight it moves. The ordering holds (Table~\ref{tab:posthoc}). Row-level restoration, which replaces some rows entirely and leaves their neighbours untouched, is the most destructive. Ablating the shared subspace, which cuts one direction out of every row, is intermediate. WiSE-FT interpolation~\citep{wortsman2022robust}, which scales every parameter by the same factor and so moves drift and compensation together, is the one edit that traces a non-catastrophic frontier. Interpolating from the retrained checkpoint back toward base at $\alpha = 0.95, 0.9, 0.8, 0.7, 0.5$ recovers 11.4\%, 22.1\%, 41.0\%, 57.0\% and 81.0\% of forgetting on Qwen/Korean while giving up $1.0\%$ to $29.9\%$ of target learning, and 14.0\% through 100.5\% on Pythia-1.4B/math for $+0.5\%$ down to $-24.3\%$. Every point on that frontier still pays, and the free point of output-scoped $\epsilon$ sits above it. Two limits keep these numbers out of the main comparison: both sweeps predate the replay-free campaign, carrying 1\% replay at a single checkpoint per cell, and the 1.4B cell injects math rather than Korean, so its small baseline (0.0364 nats) is what lets the last step read above 100\%. That last case is not a counterexample to this section, it is the case the account predicts.

\end{document}